\documentclass{article}

\usepackage{subfigure}

\usepackage[final, nonatbib]{neurips_2023}

\usepackage[colorlinks = true, citecolor = blue, urlcolor = blue, linkcolor = blue]{hyperref}
\usepackage[sort&compress]{natbib}

\usepackage{booktabs,tabularx, colortbl}
\usepackage{tabularray}
\usepackage{rotating}
\usepackage{color}
\usepackage{caption}
\usepackage{amsmath}
\usepackage{amssymb}
\usepackage[linesnumbered,ruled]{algorithm2e}
\usepackage{algpseudocode}
\usepackage{adjustbox}
\usepackage{multirow}
\usepackage{diagbox}
\usepackage{float}
\usepackage{xcolor}
\usepackage{colortbl}
\usepackage{array}
\usepackage{graphicx}
\usepackage[utf8]{inputenc} 
\usepackage[T1]{fontenc}    
\usepackage{url}            
\usepackage{booktabs}       
\usepackage{amsfonts}       
\usepackage{nicefrac}       
\usepackage{microtype}      
\usepackage{xcolor}         
\usepackage{slashbox}
\usepackage{enumerate}
\usepackage{amsthm}
\makeatletter
\def\rd@fallbacksout#1{\rd@sout#1 \rd@end}
\IfFileExists{ulem.sty}{\usepackage[normalem]{ulem}}{\let\sout\rd@fallbacksout}
\def\rd@sout#1 #2\rd@end{\rd@word{#1}\if\relax\detokenize{#2}\relax\else\rd@space\rd@sout#2\rd@end\fi}
\def\rd@word#1{\leavevmode\setbox\z@\hbox{#1}\rlap{\vrule height .62ex depth -.52ex width \wd\z@}\box\z@}
\def\rd@space{\leaders\hrule height .62ex depth -.52ex\hskip .33em plus .11em minus .07em\relax}
\makeatother

\SetAlCapNameFnt{\footnotesize}
\SetAlCapFnt{\footnotesize}
\SetAlgorithmName{Algorithm}{Algorithm}{List of Algorithms}

\title{Conformal Meta-learners for Predictive Inference of Individual Treatment Effects}

\author{%
  Ahmed M. Alaa \\
  UC Berkeley and UCSF\\
  \texttt{amalaa@berkeley.edu} \\
  \And 
  Zaid Ahmad \\
  UC Berkeley\\
  \texttt{zaidahmad@berkeley.edu} \\
  \And 
  Mark van der Laan \\
  UC Berkeley\\
  \texttt{laan@stat.berkeley.edu} \\
}

\begin{document}

\maketitle
\begin{abstract}
We investigate the problem of machine learning-based (ML)~predictive~inference~on individual treatment effects (ITEs). Previous work has focused primarily on developing ML-based ``meta-learners'' that can provide point estimates of the conditional average treatment effect (CATE)---these are model-agnostic approaches for combining intermediate nuisance estimates to produce estimates of CATE. In this paper, we develop {\it conformal meta-learners}, a general framework for issuing predictive intervals for ITEs by applying the standard conformal prediction~(CP)~procedure~on top of CATE meta-learners. We focus on a broad class of meta-learners based on two-stage pseudo-outcome regression and develop a {\it stochastic ordering} framework to study their validity. We show that~inference~with~conformal meta-learners is marginally valid if their (pseudo-outcome) conformity scores~stochastically~dominate ``oracle'' conformity scores evaluated on the unobserved~ITEs.~Additionally,~we prove that commonly used CATE meta-learners, such as the {\it doubly-robust} learner, satisfy a model- and distribution-free stochastic (or convex)~dominance~condition, making their inferences valid above a distribution-dependent coverage threshold, up to an exponentially small slack term. Whereas existing procedures~conduct~inference on nuisance parameters (i.e., potential outcomes) via weighted~CP~\cite{lei2021conformal},~conformal meta-learners enable direct inference on the target parameter (ITE). Experiments show that conformal meta-learners provide valid intervals with competitive efficiency while retaining the point estimation accuracy of CATE meta-learners.
\end{abstract}
\vspace{-0.15in}

\,\,\,\,\,\,\,\,\,\,\,\,\,\,\,\,\,\,\,\,\,{\bf Code:} \href{https://github.com/AlaaLab/conformal-metalearners}{https://github.com/AlaaLab/conformal-metalearners}

\section{Introduction}
\label{Sec1}
Identifying heterogeneity in the effects of interventions across individuals is a central problem in various fields, including medical, political, and social sciences \cite{imai2011estimation, yeager2019national, greenfield2007heterogeneity}. In recent years, there has been growing interest in developing machine learning (ML) models to estimate heterogeneous treatment effects using observational or experimental data \cite{kunzel2019metalearners,shalit2017estimating,wager2018estimation,hill2011bayesian,alaa2017bayesian}. However, most of these models only provide {\it point} estimates of the conditional average treatment effect (CATE), which is a {deterministic function that describes the expected treatment effect based on a given individual's covariates. In this paper, we focus on quantifying uncertainty in these estimates, which arises from both errors in the model and the variation of individual treatment effects (ITEs) for individuals with the same covariates. We adopt a {\it predictive inference} approach to this problem, with the goal of devising valid procedures to issue predictive intervals that cover ITEs on unseen data with a predetermined probability.

Traditionally, predictive inference on ITEs has been conducted through Bayesian methods such as BART \cite{hill2011bayesian} and Gaussian processes \cite{alaa2017bayesian}. These methods can provide interval-valued~predictions~of~ITEs through their induced posterior distributions (e.g., posterior credible intervals). However, Bayesian methods tend to be model-specific and cannot be straightforwardly generalized to modern ML models, e.g., transformer-based architectures used to model visual and textual covariate spaces \cite{li2019visualbert}. More importantly, Bayesian methods generally do not provide guarantees on the frequentist coverage of their credible intervals---achieved (finite-sample) coverage depends on the prior \cite{szabo2015frequentist}.~This~paper~is motivated by the advent of {\it conformal prediction} (CP), a frequentist alternative that can be used to conduct model-agnostic, distribution-free valid predictive inference on top~of~any~ML~model~\cite{vovk1999machine,vovk2005algorithmic, papadopoulos2008inductive}. Throughout this paper, we will study the validity of CP-based procedures for inference of ITEs.  

\textbf{\textit{What makes CP-based inference of ITEs different from its application to the standard regression (supervised) setup?}} The ``fundamental problem of causal inference''~is~that~we~never~observe~counterfactual outcomes \cite{imbens2015causal}. That is, our ``label'' is the ITE which is a difference between two potential outcomes (treated and untreated) for an individual subject---this label is never observed for any given subject because we only ever observe factual outcomes. This poses two challenges \cite{curth2021inductive}:

\textbf{(1) How to handle \textit{covariate shift}?} When treatments are assigned to individuals with probabilities that depend on their covariates, then the distributions of covariates in treated and untreated~groups~differ. Consequently, the distribution of training data differs from that of the target population.

\textbf{(2) How to incorporate \textit{inductive biases}?} Unlike supervised learning wherein we fit a single function using examples of covariates and {\it observed} targets, models of treatment effects cannot be~directly~fit~to the {\it unobserved} effects. Thus, estimates of treatment effects comprise~intermediate~estimates~of~nuisance parameters. Different approaches for combining nuisance estimates entail different inductive priors on the potential outcomes that affect the accuracy of the resulting ITE estimates. 

The literature on ML-based CATE estimation focuses on addressing the two~questions~above.~{\bf Covariate shift} affects the generalization performance of ML models---existing CATE estimation models address this problem using importance weighting \cite{shimodaira2000improving} or balanced representation learning methods for unsupervised domain adaptation \cite{ganin2015unsupervised, johansson2016learning, shalit2017estimating}. In \cite{kunzel2019metalearners}, the notion of ``meta-learners''~was~coined~to describe various model-agnostic approaches to incorporating {\bf inductive biases} and combining nuisance estimates. In \cite{kunzel2019metalearners, kennedy2020towards}, it was shown that the choice of the meta-learner influences~the~CATE~estimation rates. While the impact of {\bf (1)} and {\bf (2)} on the generalization performance of~CATE~estimators~has been extensively investigated, their impact on the validity and efficiency of predictive inference methods for ITE is less well-understood. This forms the central focus of our paper. 

{\bf Contributions.} We propose a CP procedure for predictive inference of ITEs that jointly addresses {\bf (1)} and {\bf (2)} in an end-to-end fashion. Our proposed inference strategy applies the standard CP procedure on top of a broad class of CATE meta-learners based on two-stage {\it pseudo-outcome}~regression.~These meta-learners operate by first estimating pseudo-outcomes, i.e., transformed targets~that~depend~on observable variables only, and then regressing the pseudo-outcomes on covariates to obtain point estimates of CATE. We then construct intervals for ITEs by computing the empirical quantile of conformity scores evaluated on pseudo-outcomes in a held-out calibration set. Conformal meta-learners address {\bf (1)} because the distribution of covariates associated with pseudo-outcomes~is~the~same for training and testing data, and they address {\bf (2)} since the calibration step is decoupled from model architecture, enabling flexible choice of inductive priors and the possibility of re-purposing existing meta-learners and architectures that have been shown to provide accurate estimates of CATE. 

Conformal meta-learners inherit the guarantees of CP, i.e., their resulting intervals cover pseudo-outcomes on test data with high probability. However, the original CP guarantees do not immediately translate to guarantees on coverage of ITEs. To this end, we develop a unified {\it stochastic ordering} framework to study the validity of conformal meta-learners for inference~on~ITEs.~We~show~that~inference with conformal meta-learners is valid if their conformity scores satisfy certain stochastic ordering conditions with respect to ``oracle'' conformity scores evaluated on unobserved ITEs. We prove that some of the commonly used meta-learners, such as the {\it doubly-robust} learner \cite{kennedy2020towards}, satisfy a weaker stochastic (or convex) dominance condition which makes their inferences valid above a distribution-dependent coverage threshold. Our numerical experiments show that, with careful choice of the pseudo-outcome transformation, conformal meta-learners inherit both the coverage properties of CP as well as the efficiency and point estimation accuracy of their underlying CATE meta-learners.

\section{Predictive Inference of Individual Treatment Effects (ITEs)}
\subsection{Problem setup}
\label{Sec21}
We consider the standard potential outcomes (PO) framework with a binary treatment (\cite{neyman1923application, rubin1990application}). Let $W \in \{0,1\}$ be the treatment indicator, $X \in \mathcal{X}$ be the covariates, and~$Y\in \mathbb{R}$~be~the~outcome~of~interest. For each subject $i$, let $(Y_i(0), Y_i(1))$ be the pair of potential outcomes under $W=0$ and $W=1$, respectively. The fundamental problem of causal inference is~that~we~can~only observe the {\it factual} outcome, i.e., the outcome $Y_i = W_i Y_i(1) + (1-W_i) Y_i(0)$ determined by $W_i$, but we cannot observe the {\it counterfactual} $Y_i(1-W_i)$. For $n$ subjects, we assume that the data generation process
\begin{align}
    (X_i, W_i, Y_i(0), Y_i(1)) \,\overset{iid}{\sim}\, P(X, W, Y(0), Y(1)),\, i = 1,\ldots, n,
\label{eq0}
\end{align}
satisfies the following assumptions: {\it (1) Unconfoundedness:} $(Y(0), Y(1)) \perp W \,|\, X$, {\it (2) Consistency:} $Y = Y(W)$, and {\it (3) Positivity:} $0 < P(W=1\,|\,X=x) < 1,\, \forall x \in \mathcal{X}$. These assumptions are necessary for identifying the causal effects of the treatment from a dataset $\{Z_i=(X_i, W_i, Y_i)\}_{i=1}^n$. The causal effect of the treatment on individual $i$, known as the {\it individual treatment effect} (ITE), is defined as the difference between the two potential outcomes, i.e., $Y_i(1) - Y_i(0)$.

Previous modeling efforts (e.g., \cite{hill2011bayesian, shalit2017estimating, wager2018estimation, kunzel2019metalearners}) have focused primarily on the (deterministic) {\it conditional average treatment effect} (CATE), i.e., $\tau(x) \triangleq \mathbb{E}[Y(1)-Y(0)\,|\,X=x]$. In this paper, we focus on the (random) ITE as the inferential target of interest. That is, our goal is to infer the ITE for a given subject $j$ given their covariate $X_j$ and the observed sample $\{Z_i=(X_i, W_i, Y_i)\}_{i=1}$. 

The distribution of the observed variable $Z=(X, W, Y)$ is indexed by the covariate distribution $P_X$, as well as the nuisance functions $\pi(x)$, $\mu_0(x)$ and $\mu_1(x)$ defined as follows: 
\begin{align}
\pi(x) &= \mathbb{P}(W=1\,|\,X=x),\, \nonumber \\
\mu_w(x)  &= \mathbb{E}[Y\,|\,X=x, W=w],\,\, w \in \{0, 1\}.    
\label{eq2}
\end{align}
The function $\pi(x)$ is known as the {\it propensity score} and it captures the~treatment~mechanism~underlying the data generation process. Throughout this paper, we assume that $\pi(x)$ is known, i.e.,~data~is drawn from an experimental study or the treatment assignment mechanism is known.

{\bf Predictive Inference of ITEs.} Given the sample $\{Z_i=(X_i, W_i, Y_i)\}^n_{i=1}$, our goal is to~infer~the~ITE for a new individual $n+1$ with covariate $X_{n+1}$. In particular, we would like to construct a predictive band \mbox{\small $\widehat{C}(x)$} that covers the true ITE for new test points with high probability, i.e.,
\begin{align}
\mathbb{P}(Y_{n+1}(1) - Y_{n+1}(0) \in \widehat{C}(X_{n+1})) \geq 1-\alpha,    
\label{eq3}
\end{align}
for a predetermined target coverage of $1-\alpha$, with $\alpha \in (0, 1)$, where the probability in (\ref{eq3}) accounts for the randomness of the training data $\{Z_i\}_i$ and the test point $(X_{n+1}, Y_{n+1}(1)-Y_{n+1}(0))$. Predictive intervals that~satisfy the coverage condition in (\ref{eq3}) are said to be {\it marginally valid}. 

\subsection{Conformal prediction}
\label{Sec22}
Conformal prediction (CP) is a model- and distribution-free framework for~predictive~inference~that provides (finite-sample) marginal coverage guarantees. In what follows, we describe a variant of CP, known as {\it split} (or {\it inductive}) CP \cite{vovk1999machine,vovk2005algorithmic, papadopoulos2008inductive}, for the standard regression setup.~Given~a~training~dataset $\mathcal{D} = \{(X_i, Y_i)\}_i$, the CP procedure starts by splitting $\mathcal{D}$ into two disjoint subsets: a proper training set $\{(X_j, Y_j): j \in \mathcal{D}_t\}$, and a {\it calibration} set $\{(X_k, Y_k): k \in \mathcal{D}_c\}$. Then, an ML model $\widehat{\mu}(x)$ is fit using the samples in $\mathcal{D}_t$ and a {\it conformity score} $V (.)$ is evaluated for all samples in $\mathcal{D}_c$ as follows:
\begin{equation}
V_k(\widehat{\mu}) \triangleq V(X_k, Y_k; \widehat{\mu}),\, \forall k \in \mathcal{D}_c.
\label{eq6}
\end{equation}
The conformity score measures how unusual the prediction looks relative to previous examples. A common choice of $V (.)$ is the absolute residual, i.e., $V(x, y; \widehat{\mu}) \triangleq |\,\widehat{\mu}(x) - y\,|$. For a target coverage level of $1-\alpha$, we then compute a quantile of the empirical distribution of conformity scores, i.e.,
\begin{equation}
Q_\mathcal{V}(1-\alpha) \triangleq (1-\alpha)(1+1/|\mathcal{D}_c|) \mbox{-th quantile of $\mathcal{V}(\widehat{\mu})$,}  
\label{eq7}
\end{equation}
where $\mathcal{V}(\widehat{\mu}) = \{V_k(\widehat{\mu}) : k \in \mathcal{D}_c\}$. Finally, the predictive interval at a new point $X_{n+1}=x$ is 
\begin{equation}
\widehat{C}(x) = [\,\widehat{\mu}(x) - Q_\mathcal{V}(1-\alpha),\,\widehat{\mu}(x) + Q_\mathcal{V}(1-\alpha)\,].
\label{eq8}
\end{equation}
The interval in (\ref{eq8}) is guaranteed to satisfy marginal coverage,~i.e.,~$\mathbb{P}(Y_{n+1}\in\widehat{C}(X_{n+1})) \geq 1-\alpha$.~The only assumption needed for this condition to hold is the {\it exchangeability} between calibration and test data \cite{vovk1999machine, lei2013distribution, lei2018distribution}. Note that the interval in (\ref{eq8}) has a fixed length of $2 Q_\mathcal{V}(1-\alpha)$~that~is~independent~of~$x$. To enable adaptive intervals, \cite{romano2019conformalized} proposed a variant of the CP procedure where the base model is a quantile regression with interval-valued predictions $[\widehat{\mu}_{\alpha/2}(x), \widehat{\mu}_{1-\alpha/2}(x)]$,~and~the~conformity~score~is defined as the signed distance $V_k(\widehat{\mu}) \triangleq \max\{\widehat{\mu}_{\alpha/2}(X_k)-Y_k, Y_k -\widehat{\mu}_{1-\alpha/2}(X_k)\}$. 

\subsection{Oracle conformal prediction of ITEs}
\label{Sec23} 
How can we adapt the CP framework for predictive inference of ITEs? In a hypothetical world where we have access to counterfactual outcomes, we can apply the standard CP in~Section~\ref{Sec22}~to~a~dataset~of covariates and ITE tuples, $\mathcal{D}^*=\{(X_i, Y_i(1) - Y_i(0))\}_i$, and compute conformity scores as: 
\begin{equation}
V^*_k(\widehat{\tau}) \triangleq V(X_k, \colorbox{red!10}{$\textcolor{black}{Y_k(1)}$}-\colorbox{blue!5}{$\textcolor{black}{Y_k(0)}$}; \widehat{\tau}),\, \forall k \in \mathcal{D}^*_c,
\label{eq8b}
\end{equation}
where $\widehat{\tau}$ is an ML model fit to estimate the CATE function $\tau(x)$ using $\mathcal{D}^*_t$, and $\mathcal{D}^* = \mathcal{D}^*_t \cup \mathcal{D}^*_c$. We will refer to this procedure as ``oracle'' conformal prediction and to $V^*_k(\widehat{\tau})$ as the oracle conformity scores. Since the oracle problem is a standard regression, the oracle procedure is marginally valid---i.e., it satisfies the guarantee in (\ref{eq3}), $\mathbb{P}(Y(1)-Y(0) \in \widehat{C}^*(X)) \geq 1-\alpha$. However, oracle CP is infeasible since we can only observe one of the potential outcomes (colored in red and~blue~in~(\ref{eq8b})),~hence we need an alternative procedure that operates only on the observed variable $Z=(X, W, Y)$.

\subsection{The two challenges of predictive inference on ITEs} 
\label{Sec24} 
A na\"ive approach to inference of ITEs is to split the observed~sample~\mbox{\small $\{Z_i=(X_i, W_i, Y_i)\}_i$}~by~the treatment group and create two datasets: \mbox{\small $\mathcal{D}_0 = \{(X_i, Y_i): \colorbox{blue!5}{$\textcolor{black}{W_i=0}$}\}_i$}, \mbox{\small $\mathcal{D}_1 = \{(X_i, Y_i): \colorbox{red!10}{$\textcolor{black}{W_i=1}$}\}_i$},~then generate two sets of conformity scores for the nuisance estimates $\widehat{\mu}_0$ and $\widehat{\mu}_1$ as follows:
\begin{equation} 
\colorbox{blue!5}{$V^{(0)}_k(\widehat{\mu}_0) \triangleq V(X_k, Y_k(0); \widehat{\mu}_0),\, \forall k \in \mathcal{D}_{c,0},$}\,\,\,\,\,\,\,\, \colorbox{red!10}{$V^{(1)}_k(\widehat{\mu}_1) \triangleq V(X_k, Y_k(1); \widehat{\mu}_1),\, \forall k \in \mathcal{D}_{c,1},$}
\label{eq8c}
\end{equation}
where $\mathcal{D}_{c,0}$ and $\mathcal{D}_{c,1}$ are calibration subsets of $\mathcal{D}_{0}$ and $\mathcal{D}_{1}$. In order to construct valid~predictive~intervals for ITE using the conformity scores in (\ref{eq8c}), we need to reconsider~how~the~two~distinct~characteristics of CATE estimation, previously discussed in Section \ref{Sec1}, interact with the CP procedure:

\begin{minipage}{.5\textwidth}
\begin{center} 
\textbf{\textit{(1) Covariate shift}}
\end{center}

The distributions of covariates for treated and untreated subjects differ from that of the target population: \mbox{\small $P_{X|W=0} \neq P_{X|W=1} \neq P_{X}$}, i.e., the following holds for the conformity scores in (\ref{eq8c}): 
\vspace{-0.05in}
\begin{equation}
\mbox{\small $\colorbox{blue!5}{$P_{X, V^{(0)}|W=0} \neq P_{X, V^{(0)}}$}\,\, \colorbox{red!10}{$P_{X, V^{(1)}|W=1} \neq P_{X, V^{(1)}}$}$}\nonumber
\end{equation} 

\end{minipage}
\enspace 
\begin{minipage}{.475\textwidth}
\begin{center}
\textbf{\textit{(2) Inductive biases}}
\end{center}

Different choices for the joint model of $\mu_0$ and $\mu_1$ encode different inductive biases that impose different forms of regularization on the implied CATE function, i.e., \mbox{\small $\widehat{\tau}(x) = \widehat{\mu}_1(x)-\widehat{\mu}_0(x)$} \cite{kunzel2019metalearners}. These biases influence the induced distributions \mbox{\small $P_{V^{(0)}}$} and \mbox{\small $P_{V^{(1)}}$} of the conformity scores in (\ref{eq8c}).
\end{minipage}
\vspace{0.01in}

Covariate shift breaks the exchangeability assumption necessary for the validity of~CP.~Current~methods have primarily focused on {\bf (1)} with $Y(0)$ and $Y(1)$ as inference targets, and developed approaches for handling covariate shift by reweighting conformity scores \cite{tibshirani2019conformal, lei2021conformal}. The resulting intervals for POs are then combined to produce intervals for ITEs. However, these method tie the CP procedure to model architecture, requiring inference on nuisance parameters, and hence lose the desirable post-hoc nature of CP. Furthermore, inference on POs is likely to provide conservative ITE intervals, and limits the inductive priors that can be assumed since not all CATE models provide explicit PO estimates.

\section{Conformal Meta-learners} 
\label{Sec3} 
In \cite{kunzel2019metalearners}, a taxonomy of ``meta-learners'' was introduced to categorize different inductive priors that can be incorporated into CATE estimators by structuring the regression models for $\mu_0$ and $\mu_1$. For example, the {\it T-learner} estimates $\widehat{\mu}_0$ and $\widehat{\mu}_1$ independently using $\mathcal{D}_0$ and $\mathcal{D}_1$, while the {\it S-learner} models the treatment variable $W$ as an additional covariate in a joint regression model $\widehat{\mu}(X, W)$ and estimates CATE as $\widehat{\tau}(x) = \widehat{\mu}(x, 1)-\widehat{\mu}(x, 0)$. In this Section, we propose an~end-to-end~solution~to~{\bf (1)} and {\bf (2)} by applying CP on top of CATE meta-learners in a post-hoc fashion, thereby decoupling the CP procedure from the CATE model and allowing direct inference on ITEs. In the next Section, we develop a unified framework for analyzing the validity of this broad class of procedures.

\begin{figure}[!t]
\begin{minipage}{.575\textwidth}
\centering
\includegraphics[width=3in]{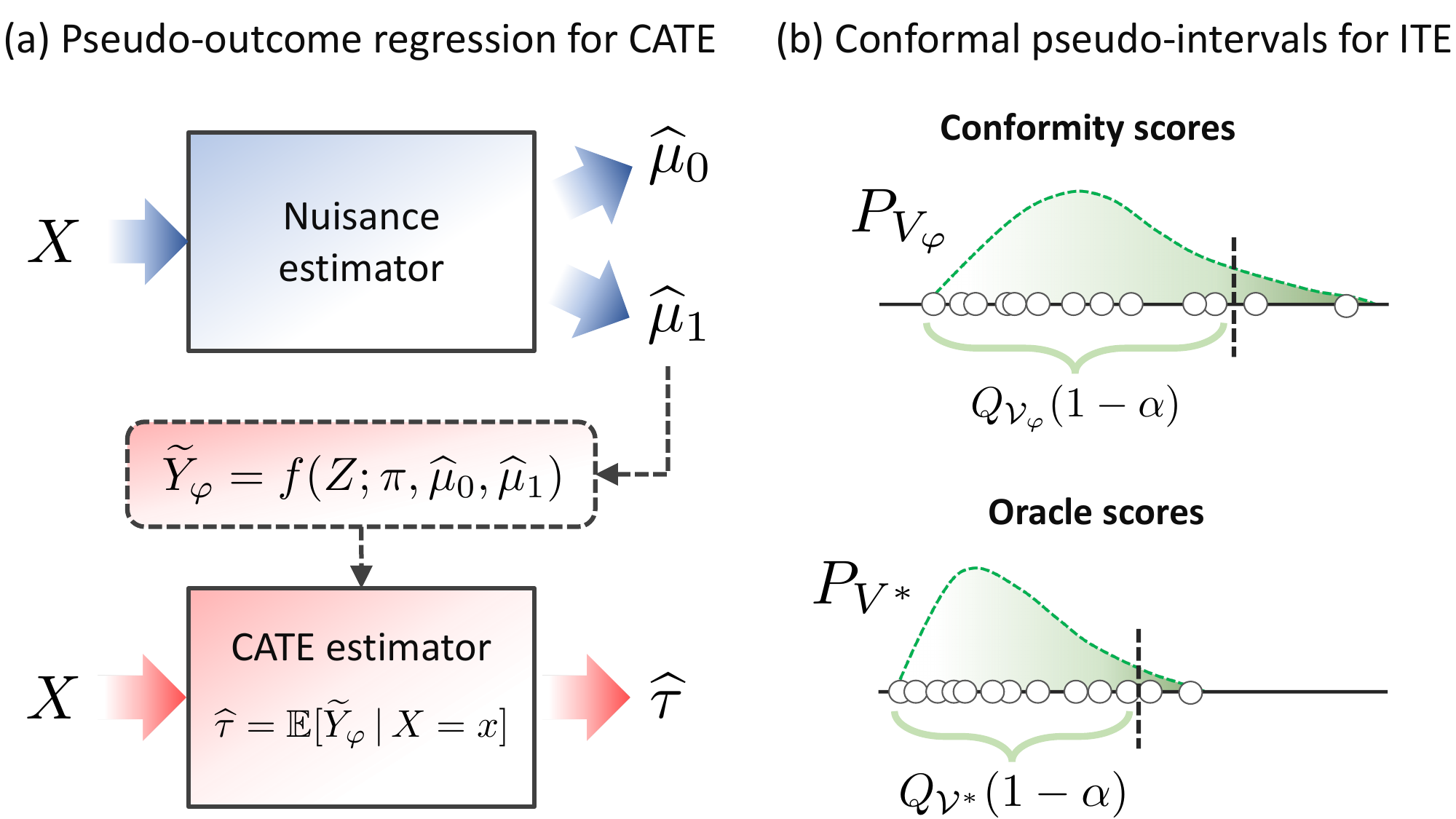}
{\footnotesize Figure 1: Pictorial depiction of conformal meta-learners.}
\end{minipage}
\quad
\begin{minipage}{.4\textwidth}
\begin{algorithm}[H]
{\footnotesize
\vspace{.025in}
    \SetKwInOut{Input}{{\footnotesize Input}}
    \SetKwInOut{Output}{{\footnotesize Output}}
    {\color{black}
    \Input{\, \mbox{\scriptsize $\mathcal{D} = \{(X_i, W_i, Y_i)\}_{i=1}^n$}, \mbox{\scriptsize $X_{n+1}$}}
    \vspace{.01in}	
    \Output{\, \mbox{\scriptsize $\widehat{C}_\varphi(X_{n+1})$}}}
    \vspace{.03in}
    \vspace{.05in}
    
	{\color{black}
	{\it Split \mbox{\scriptsize $\mathcal{D}$} into \mbox{\scriptsize $\mathcal{D}_{\varphi}$}, \mbox{\scriptsize $\mathcal{D}_{t}$} and \mbox{\scriptsize $\mathcal{D}_{c}$}}}\;
	\vspace{.025in}
	\textbf{Pseudo-outcome regression:}
 
	\vspace{.03in}
	{\bf (1)} Estimate \mbox{\scriptsize $\widehat{\varphi}=(\pi, \widehat{\mu}_0, \widehat{\mu}_1)$} using \mbox{\scriptsize $\mathcal{D}_{\varphi}$}\; 
	{\bf (2)} Regress \mbox{\scriptsize $\widetilde{Y}_{\varphi} = f(Z, \widehat{\varphi})$}~on~\mbox{\scriptsize $X$}~using~\mbox{\scriptsize $\mathcal{D}_{t}$}\; 
	\vspace{.05in}	
	\textbf{Conformal pseudo-intervals:}
 
	\vspace{.03in}
	Evaluate conformity scores \mbox{\scriptsize $\mathcal{V}_\varphi$} using \mbox{\scriptsize $\mathcal{D}_{c}$}\; 
	\vspace{.025in}	
	{\bf Return} \mbox{\scriptsize $\widehat{C}_\varphi = [\widehat{\tau}(X_{n+1}) \pm Q_{\mathcal{V}_\varphi}(1-\alpha)]$}
	\vspace{.015in}	}
 \caption{{\footnotesize {\bf Conformal Meta-learner}}}
\end{algorithm}
\end{minipage}
\arrayrulecolor{lightgray}
\vspace{-.15in}
\end{figure}

\subsection{Pseudo-outcome regression for CATE estimation} 
\label{Sec31} 
We focus on a broad subclass of CATE meta-learners based on two-stage {\it pseudo-outcome} regression. These models replace the (unobserved) oracle ITEs with ``proximal'' targets that~are~estimated~from~observed variables only, and then train an ML model to predict the estimated~targets~from~covariates.~The two stages of this general pseudo-outcome regression framework can be described as follows:

{\bf Stage 1.} We obtain a plug-in estimate \mbox{\small $\widehat{\varphi}$} of the nuisance parameters~\mbox{\small $\varphi = (\pi, \mu_0, \mu_1)$}. Note that~since~we assume that the propensity score is known, we only need to estimate \mbox{\small $\mu_0$} and \mbox{\small $\mu_1$} using \mbox{\small $\mathcal{D}_0$} and \mbox{\small $\mathcal{D}_1$}.

{\bf Stage 2.} We use the nuisance estimates obtained in Stage 1 to create pseudo-outcomes \mbox{\small $\widetilde{Y}_{\varphi}$}~that~depend only on \mbox{\small $\widehat{\varphi}$} and the observable variables \mbox{\small $Z=(X, W, Y)$}, i.e., \mbox{\small $\widetilde{Y}_{\varphi} = f(Z, \widehat{\varphi})$}~for~some~function~\mbox{\small $f$}. The CATE estimate is then obtained by regressing the pseudo-outcome \mbox{\small $\widetilde{Y}_{\varphi}$} on the~covariate~\mbox{\small $X$}.~This~is~typically conducted using a different dataset than the one used to obtain the nuisance estimate \mbox{\small $\widehat{\varphi}$}.   

\begin{minipage}{.275\textwidth}
The general framework described above captures various models~in~previous~literature. We~study~3~examples of such meta-learners: X-learner, Inverse propensity weighted (IPW) learner and doubly-robust (DR) learner. 

\end{minipage}
\enspace 
\begin{minipage}{.725\textwidth}
\centering
{\footnotesize
\begin{tabular}{rllll}
\specialrule{.2em}{.1em}{.1em}  \\[-4ex] 
& \\ 
\multicolumn{1}{l}{} & \multicolumn{4}{c}{{\bf Pseudo-outcome}}                                                                                                               \\ [0.5ex] \hline 
 &   \\[-1.5ex]
{\it IPW-learner} \cite{horvitz1952generalization} & \multicolumn{4}{l}{\hspace{-1em} $\widetilde{Y}_{\varphi} =  \frac{W-\pi(X)}{\pi(X)(1-\pi(X))} Y$}                                                                     \\[5pt] 
{\it X-learner} \cite{kunzel2019metalearners} & \multicolumn{4}{l}{\hspace{-1em} $\widetilde{Y}_{\varphi} = W(Y-\widehat{\mu}_0(X)) + (1-W)(\widehat{\mu}_1(X)-Y)$}                                               \\[5pt] %
{\it DR-learner} \cite{kennedy2020towards} & \multicolumn{4}{l}{\hspace{-1em} $\widetilde{Y}_{\varphi} =  \frac{W-\pi(X)}{\pi(X)(1-\pi(X))} (Y-\widehat{\mu}_W(X)) + \widehat{\mu}_1(X) - \widehat{\mu}_0(X)$} \\ [5pt] 
\specialrule{.2em}{.1em}{.1em} 
\end{tabular}} 
\vspace{.05in}

{\footnotesize Table 1: Existing meta-learners as instantiations of pseudo-outcome regression.}
\end{minipage}

Table 1 lists the pseudo-outcomes \mbox{\small $\widetilde{Y}_{\varphi}$} for the three meta-learners: IPW-learner reweights factual outcomes using propensity scores to match CATE, i.e., \mbox{\small $\mathbb{E}[\widetilde{Y}_{\varphi}\,|\,X=x] = \tau(x)$}; X-learner uses regression adjustment to impute counterfactuals; DR-learner combines both approaches. DR- and X-learners\footnote{Here, we consider a special case of the X-learner in \cite{kunzel2019metalearners} which involves a weighted sum of two regression adjusted models \mbox{\small $\widehat{\tau}_0$} and \mbox{\small $\widehat{\tau}_1$} trained separately on the treated and control datasets \mbox{\small $\mathcal{D}_0$} and \mbox{\small $\mathcal{D}_1$}.}, coupled with specific architectures for joint modeling of \mbox{\small $\widehat{\mu}_0$} and \mbox{\small $\widehat{\mu}_1$}, have shown competitive performance for CATE estimation in previous studies \cite{kunzel2019metalearners, curth2021inductive, kennedy2020towards}. The conformal meta-learner framework decouples the CP procedure (Section \ref{Sec32}) from the inductive priors encoded by these meta-learners, hence it inherits their favorable CATE estimation properties and enables a potentially more efficient direct inference on ITEs as opposed to inference on POs. This addresses {\bf challenge (2)} in Section \ref{Sec24}.

\subsection{Conformal pseudo-intervals for ITEs} 
\label{Sec32} 
Pseudo-outcome regression is based on the notion that accurate proxies for~treatment~effects~can~produce reliable CATE point estimates. This concept can be extended to predictive inference:~using~CP~to calibrate meta-learners via held-out pseudo-outcomes can yield accurate ``pseudo-intervals'' for ITEs.

Given a dataset \mbox{\small $\mathcal{D} = \{Z_i=(X_i, W_i, Y_i)\}_i$}, we create three mutually-exclusive subsets: \mbox{\small $\mathcal{D}_{\varphi}$}, \mbox{\small $\mathcal{D}_t$} and \mbox{\small $\mathcal{D}_c$}. \mbox{\small $\mathcal{D}_{\varphi}$} is used to estimate the nuisance parameters \mbox{\small $\varphi$}. Next, the estimates \mbox{\small $\widehat{\varphi}=(\pi, \widehat{\mu}_0, \widehat{\mu}_1)$} are~used~to~transform \mbox{\small $\{Z_i=(X_i, W_i, Y_i): i \in \mathcal{D}_t\}$} into covariate/pseudo-outcome pairs \mbox{\small $\{(X_i, \widetilde{Y}_{\varphi, i}): i \in \mathcal{D}_t\}$}~which~are used to train a CATE model \mbox{\small $\widehat{\tau}$}. Finally, we compute conformity scores for \mbox{\small $\widehat{\tau}$} on pseudo-outcomes, i.e.,
\begin{equation}
V_{\varphi, k}(\widehat{\tau}) \triangleq V(X_k, \widetilde{Y}_{\varphi, k}; \widehat{\tau}),\, \forall k \in \mathcal{D}_c.
\label{eq10}
\end{equation}
For a target coverage of $1-\alpha$, we construct a predictive interval at a new point $X_{n+1}=x$ as follows: 
\begin{equation}
\widehat{C}_\varphi(x) = [\,\widehat{\tau}(x) - Q_{\mathcal{V}_\varphi}(1-\alpha),\,\widehat{\tau}(x) + Q_{\mathcal{V}_\varphi}(1-\alpha)\,],
\label{eq10b}
\end{equation}
where $\mathcal{V}_\varphi = \{V_{\varphi, k}(\widehat{\tau}): k \in \mathcal{D}_c\}$. We call $\widehat{C}_\varphi(x)$ a {\it pseudo-interval}.~The~conformal~meta-learner~approach is depicted in Figure 1 and a summary of the procedure is given in Algorithm 1.

Note that conditional on~$\widehat{\varphi}$, the pseudo-outcomes \mbox{\small $(X, \widetilde{Y}_{\varphi})$} in calibration data are~drawn~from~the~target distribution, which maintains the exchangeability of conformity scores and addresses covariate shift ({\bf challenge (1)} in Section \ref{Sec24}). However, the conformity scores \mbox{\small $V{\varphi}$} are evaluated on transformed outcomes, which means that \mbox{\small $V_{\varphi}$} and \mbox{\small $V^*$} are not exchangeable, even though they are drawn from the same covariate distribution. Consequently, the usual CP guarantees, i.e., \mbox{\small $\mathbb{P}(\widetilde{Y}_{\varphi}\in\widehat{C}_\varphi(X)) \geq 1-\alpha$}, do not immediately translate to coverage guarantees for the true ITE \mbox{\small $Y(1)-Y(0)$}. In the next section, we show that for certain choices of the pseudo-outcomes, the corresponding pseudo-intervals can provide valid inferences for ITE without requiring the exchangeability of \mbox{\small $V_{\varphi}$} and \mbox{\small $V^*$}.

\begin{figure*}[t]
    \centering
    \includegraphics[width=5.5in]{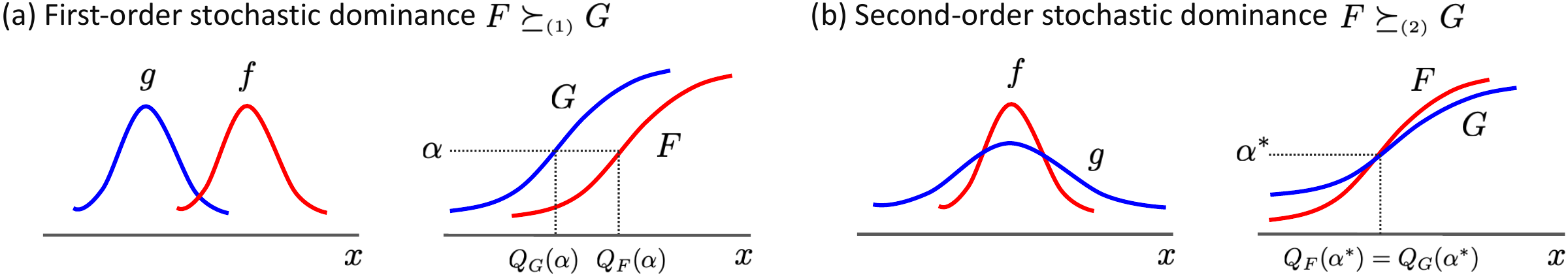}
    \caption*{{\footnotesize Figure 2: Graphical illustration of stochastic dominance among two exemplary distributions~\mbox{\footnotesize $F$}~and~\mbox{\footnotesize $G$}.}}
    \vspace{-0.2in}
\end{figure*}

\section{Validity of Conformal Meta-learners: A Stochastic Ordering Framework} 
\label{Sec4} 
Under what conditions are pseudo-intervals valid for inference of ITEs? Recall that these~intervals~are constructed by evaluating the empirical quantile of pseudo-outcome~conformity~scores.~Intuitively,~the pseudo-intervals will cover the true ITE if the conformity scores are ``stochastically larger'' than the oracle scores in Section \ref{Sec23}, i.e., $Q_{\mathcal{V}_\varphi}(\alpha) \geq Q_{\mathcal{V}^*}(\alpha)$ in some stochastic sense~(Figure~1(b)).~Hence, to study the validity of conformal meta-learners, we analyze the {\it stochastic orders} of $V_\varphi$ and $V^*$, and identify conditions under which pseudo-intervals cover oracle intervals. 

Stochastic orders are partial orders of random variables used to compare their location, magnitude, or variability \cite{shaked2007stochastic, rothschild1978increasing}. In our analysis, we utilize two key notions of stochastic order~among~cumulative distribution functions (CDFs) $F$ and $G$, which we formally define below. 

{\bf Definition 1} \textbf{(Stochastic dominance)} $F$ has {\it first-order} stochastic dominance (FOSD)~on~$G$,~$F \succeq_{\mbox{\tiny (1)}}G$, iff $F(x) \leq G(x), \forall x$, with strict inequality for some $x$. $F$ has {\it second-order}~stochastic~dominance (SOSD) over $G$, $F \succeq_{\mbox{\tiny (2)}}G$, iff \mbox{\small $\int_{-\infty}^{x} \left[G(t)-F(t)\right]dt \geq 0$}, $\forall x$, with strict inequality~for~some~$x$.  

{\bf Definition 2} \textbf{(Convex dominance)} $F$ has monotone {\it convex} dominance (MCX) over $G$, $F \succeq_{mcx}G$, iff \mbox{\small $\mathbb{E}_{X\sim F}[u(X)] \geq \mathbb{E}_{X\sim G}[u(X)]$} for all non-decreasing convex functions $u: \mathbb{R}\to\mathbb{R}$.

Stochastic ordering is useful tool in decision theory and quantitative finance~used~to~analyze~the~decisions of utility maximizers with varying risk attitudes \cite{asmussen2020risk}. A distribution $F$ has FOSD over~$G$~if~it~is favored by any decision-maker with a non-decreasing utility function, i.e., $F$ is more likely to give higher outcomes than $G$ because its CDF is strictly lower (Figure 2(a)). If $F$ has SOSD over $G$, then it is favored by risk-averse decision-makers, i.e., $f$ has smaller spread than $g$ and is favored by all decision-makers with a non-decreasing {\it concave} utility function \cite{hadar1969rules}. In this case, the CDFs can cross but $G$ is always higher before the first crossing point (Figure 2(b)). $F$ has MCX over $G$ if it is favored by decision-makers with a non-decreasing {\it convex} utility---in this case, the CDFs can cross finitely many times, but $F$ is always lower after the last crossing point (See Appendix A for more details). 

In the following Theorem, we provide sufficient conditions for the validity of conformal meta-learners in terms of the stochastic orders of their conformity scores.

{\bf Theorem 1.} {\it If $(X_i, W_i, Y_i(0), Y_i(1)),\, i=1,\ldots, n+1$ are i.i.d., then $\exists\, \alpha^* \in (0, 1]$ such that the pseudo-interval $\widehat{C}_\varphi(X_{n+1})$ constructed using the dataset $\mathcal{D}=\{(X_i, W_i, Y_i)\}_{i=1}^n$ satisfies} 
\begin{align}
\mathbb{P}(Y_{n+1}(1)-Y_{n+1}(0)\in\widehat{C}_\varphi(X_{n+1})\,|\,\mathcal{D}_\varphi\cup\mathcal{D}_t) \geq 1-\alpha-\epsilon_{n_c}(\alpha),\, \forall \alpha \in (0, \alpha^*), \nonumber
\end{align}
{\it if at least one of the following stochastic ordering conditions hold for the distributions of $V_\varphi$ and $V^*$ conditional on $\mathcal{D}_\varphi\cup\mathcal{D}_t$: (i) $V_\varphi \succeq_{\mbox{\tiny (1)}} V^*$,~or (ii) $V_\varphi \succeq_{mcx} V^*$ and $F_{V_\varphi}-F_{V^*}$ changes sign finitely many times. Under (i), we have $\alpha^*=1$ and $\epsilon_{n_c}(\alpha)=0$.~Under~(ii), $\epsilon_{n_c}(\alpha)=\exp(-2n_c(\alpha^*-\alpha)^2)$, where $n_c=|\mathcal{D}_c|$ and $\alpha^*=1-F_{V_\varphi}(v^*)$, with $v^*$ the last crossing point of $F_{V_\varphi}$ and $F_{V^*}$.} 

All proofs are provided in Appendix A. Theorem 1 states that if the conformity~score~$V_\varphi$~of~the~meta-learner is stochastically larger (FOSD) or has a larger spread (MCX)~than~the~oracle~conformity score, then the conformal meta-learner is valid for high-probability coverage (Figure 3), up to a slack term that vanishes exponentially fast in sample size under MCX. (This is the range of target coverage that is of practical relevance, i.e., $\alpha$ is typically set to 0.05 or 0.1. Note, however, that under MCX the threshold $\alpha^*$ depends on the distribution of the conformity scores and can be arbitrarily small, so validity at a fixed level such as $\alpha=0.1$ is not guaranteed; see the Remark in Appendix A.3.) Because stochastic (or convex) dominance pertain to more variable conformity scores, the predictive intervals of conformal meta-learners will naturally be more conservative than the oracle intervals for $\alpha<\alpha^*$.

\begin{figure}[!t]
\begin{minipage}{.4\textwidth}
\centering
\includegraphics[width=2.1in]{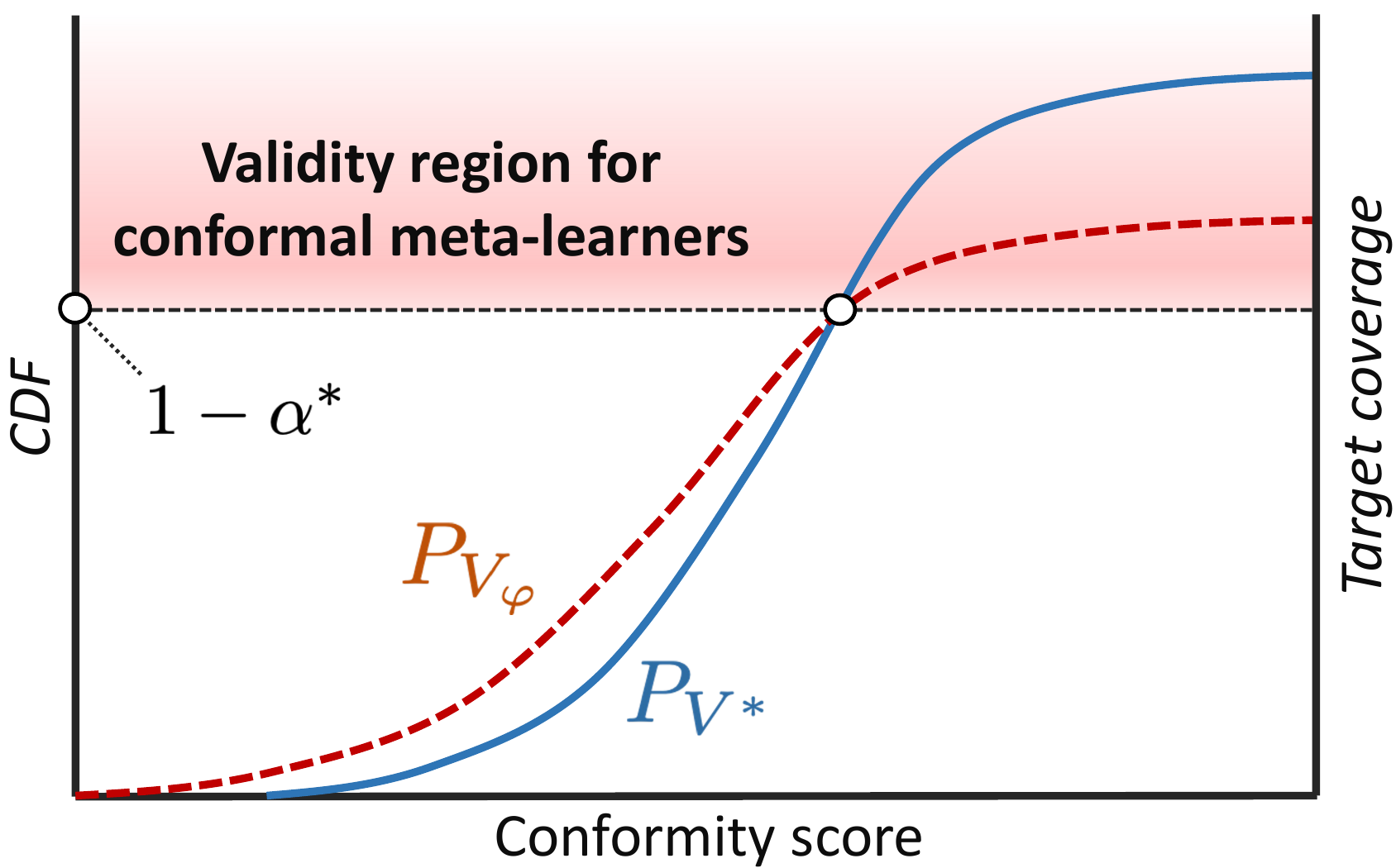}
{\footnotesize Figure 3: Validity condition in Theorem 1.}
\end{minipage}
\enspace 
\begin{minipage}{.6\textwidth}
\centering
{\footnotesize
\begin{tabular}{r|clcl}
\specialrule{.2em}{.1em}{.1em}  \\[-2ex]
\multicolumn{1}{l|}{\multirow{2}{*}{\textbf{Meta-learner}}} & \multicolumn{4}{c}{\textbf{Conformity score}}                                         \\  [0.5ex] 
\multicolumn{1}{l|}{}                  & \multicolumn{2}{c}{\textit{Absolute residual}} & \multicolumn{2}{c}{\textit{Signed distance}~\cite{linusson2014signed}} \\ [0.5ex] \cline{1-5} 
& & \\ [-1.5ex] 
\textit{X-learner}                     & \multicolumn{2}{c}{No stochastic order}        & \multicolumn{2}{c}{No stochastic order}      \\[.5ex]
\textit{IPW-learner}                   & \multicolumn{2}{c}{\mbox{\small $V_\varphi \succeq_{mcx} V^*$}}                                                                      & \multicolumn{2}{c}{\mbox{\small $V_\varphi 	\preceq_{\mbox{\tiny (2)}} V^*$}}                                                                    \\[.5ex]
\textit{DR-learner}                    & \multicolumn{2}{c}{\mbox{\small $V_\varphi \succeq_{mcx} V^*$}}                                                                      & \multicolumn{2}{c}{\mbox{\small $V_\varphi 	\preceq_{\mbox{\tiny (2)}} V^*$}}                                                                    \\ [.5ex]
\specialrule{.2em}{.1em}{.1em} 
\end{tabular}} 
\vspace{.05in}

{\footnotesize Table 2: Stochastic orders of conformity scores for the three meta-learners considered in Table 1.}
\end{minipage}
\vspace{-.2in}
\end{figure}

Whether a meta-learner meets conditions {\it (i) or (ii)} of Theorem 1 depends on how the pseudo-outcome, \mbox{\small $\widetilde{Y}_\varphi = f(Z; \widehat{\varphi})$}, is constructed. The following Theorem provides an answer to the question of which of the meta-learners listed in Table 1 satisfy the stochastic ordering conditions in Theorem 1.

{\bf Theorem 2.} {\it Let $V_{\varphi}(\widehat{\tau}) = |\widehat{\tau}(X)-\widetilde{Y}_{\varphi}|$ and assume that the propensity score function $\pi: \mathcal{X} \to [0, 1]$ is known. Then, the following holds: (i) For the $X$-learner, $V_\varphi$ and $V^*$ do not admit to a model- and distribution-free stochastic order, (ii) For any distribution $P(X, W, Y(0), Y(1))$, CATE estimate $\widehat{\tau}$, and nuisance estimate $\widehat{\varphi}$, the IPW- and the DR-learners satisfy $V_\varphi \succeq_{mcx} V^*$.}

Theorem 2 states that the stochastic ordering of \mbox{\small $V_\varphi$} and \mbox{\small $V^*$} depends on the specific choice of~the~conformity score function \mbox{\small $V(X, \widetilde{Y}_{\varphi}; \widehat{\tau})$} as well as the choice of the meta-learner, i.e., the pseudo-outcome generation function \mbox{\small $\widetilde{Y}_{\varphi} = f(Z; \widehat{\varphi})$}. The IPW- and DR-learners ensure that, by construction,~the~pseudo-outcome is equal to CATE in expectation: \mbox{\small $\mathbb{E}[\widetilde{Y}_{\varphi}\,|\,X=x] = \tau(x)$}. This construction enables the IPW- and DR-learners to provide unbiased estimates of average treatment effects (ATE) independent of the data distribution and the ML model used for the nuisance estimates \mbox{\small $\widehat{\mu}_0$} and \mbox{\small $\widehat{\mu}_1$}. By the~same~logic, IPW- and DR-learners also guarantee stochastic (convex) dominance of their~conformity~scores~irrespective of the data distribution and the ML model choice, so that the convex dominance in condition (ii) of Theorem 1 holds in a model- and distribution-free manner (although the threshold $\alpha^*$ remains distribution-dependent). Contrarily, the X-learner does not use the knowledge~of~$\pi$~to construct its pseudo-outcomes, hence it does not guarantee a (distribution-free) stochastic order~and the achieved coverage depends on the nuisance estimates \mbox{\small $\widehat{\mu}_0$} and \mbox{\small $\widehat{\mu}_1$}. In Table 2, we list the stochastic orders achieved for different choices of meta-learners and conformity~scores.~(The~analysis~of~stochastic orders for the signed distance score used in \cite{romano2019conformalized} and \cite{linusson2014signed} is provided in Appendix A.) 

{\bf Key limitations of conformal meta-learners.} While conformalized meta-learners can enable valid end-to-end predictive inference of ITEs, they have two key limitations. First, the propensity score $\pi$ must be known to guarantee model- and distribution-free stochastic ordering of conformity scores. However, we note that this limitation in not unique to our method and is~also~encountered~in~methods based on weighted CP \cite{tibshirani2019conformal,lei2021conformal}. The second limitation is peculiar to our method: exact characterization of \mbox{\small $\alpha^*$} is difficult and depends on the data distribution; in particular, convex dominance alone does not bound $\alpha^*$ away from zero (Appendix A.3). Devising procedures for inferring \mbox{\small $\alpha^*$} based on observable variables or deriving theoretical lower bounds on \mbox{\small $\alpha^*$} are interesting directions for future work; alternatively, the superquantile calibration in Theorem 3 removes the dependence on $\alpha^*$, at the cost of wider intervals. Here, we focus on empirical evaluation of \mbox{\small $\alpha^*$} in semi-synthetic experiments. 

\section{Experiments}
\label{Sec5}
\subsection{Experimental setup}
\label{Sec51}
Since the true ITEs are never observed in real-world datasets, we follow the common practice of conducting numerical experiments using synthetic and semi-synthetic datasets \cite{hill2011bayesian, johansson2016learning, lei2021conformal}.

\textbf{Synthetic datasets.} We consider a variant of the data-generation process in Section 3.6 in~\cite{lei2021conformal}~which was originally proposed in~\cite{wager2018estimation}. We create synthetic datasets by sampling covariates \mbox{\small $X \sim U([0,1]^d)$} with \mbox{\small $d = 10$} and treatments \mbox{\small $W \mid X = x \sim \mathrm{Bern}(\pi(x))$} with \mbox{\small $\pi(x) = (1 + I_{x_1}(2,4))/4$}, where \mbox{\small $I_{x_1}(2,4)$} is the regularized incomplete beta function (i.e., the CDF of a Beta distribution with shape parameters 2 and 4) evaluated at \mbox{\small $x_1$}. Outcomes are modeled as \mbox{\small $\mu_1(x) = \zeta(x_1) \cdot \zeta(x_2)$} and \mbox{\small $\mu_0(x) = \gamma\, \zeta(x_1) \cdot \zeta(x_2)$}, where \mbox{\small $\gamma \in [0,1]$} is a parameter that controls the magnitude of the treatment effect, and \mbox{\small $\zeta$} is a function \mbox{\small $\zeta(x) = 2/(1 + \exp(-12(x - 0.5)))$}. We sample the POs independently as \mbox{\small $Y(0) \mid X = x \sim \mathcal{N}(\mu_0(x), 1)$} and \mbox{\small $Y(1) \mid X = x \sim \mathcal{N}(\mu_1(x), \sigma^2(x))$}, with a heteroscedastic noise model \mbox{\small $\sigma^2(x) = -\log(x_1)$}. We define two setups within this model: \textbf{Setup A}, where the treatment has no effect on average (\mbox{\small $\gamma = 1$}, so the CATE is zero), and \textbf{Setup B}, where the effects are heterogeneous (\mbox{\small $\gamma = 0$}).

\begin{figure*}[t]
\centering
\includegraphics[width=5.5in]{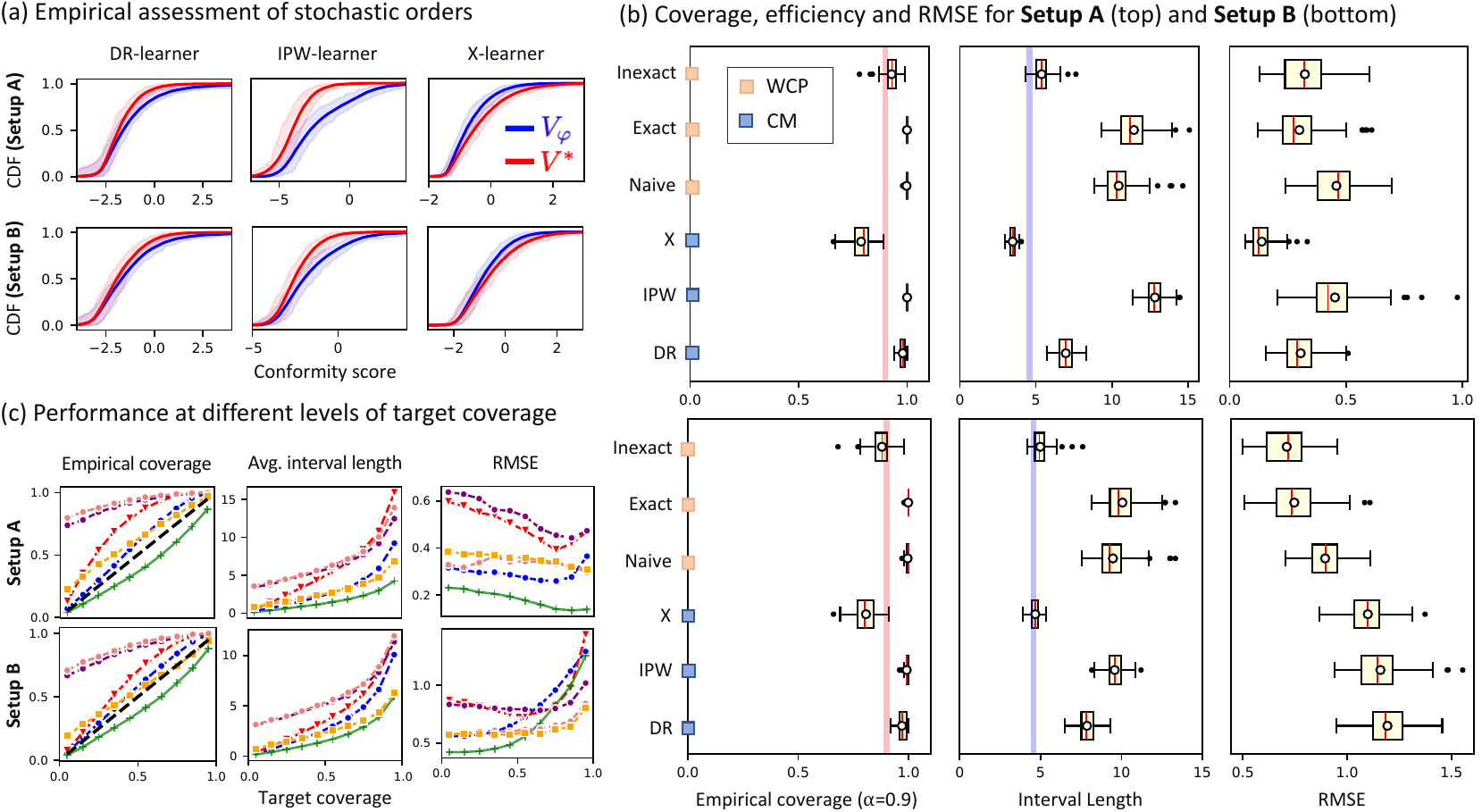}
\caption*{\footnotesize Figure 4: Performance of all baseline in the synthetic setup described in Section \ref{Sec51}. In (b), red vertical lines correspond to target coverage (\mbox{\footnotesize $1-\alpha=0.9$}), and blue vertical lines correspond to optimal interval width. In (c), baseline methods are color-coded as follows: {\color{blue}{$\bullet$}} CM-DR, {\color{red}{$\bullet$}} CM-IPW, {\color{green}{$\bullet$}} CM-X, {\color{purple}{$\bullet$}} WCP-Na\"ive, {\color{pink}{$\bullet$}} WCP-Exact, and {\color{orange}{$\bullet$}} WCP-Inexact. Here, WCP stands for weighted CP and CM stands for conformal meta-learners.}
\vspace{-.05in} 
  \rule{\linewidth}{.25pt}   
  \vspace{-.35in} 
\end{figure*}

{\bf Semi-synthetic datasets.} We also consider two well-known semi-synthetic datasets that involve real covariates, or emulations of real covariates, and simulated outcomes. The first is the National Study of Learning Mindsets (NLSM) \cite{yeager2019national, lei2021conformal}, and the second is the IHDP benchmark originally developed in \cite{hill2011bayesian}. Details on the data generation process for NLSM can be found in Section 4.4 in \cite{lei2021conformal}. Details on the IHDP benchmark can be found in \cite{hill2011bayesian, johansson2016learning, shalit2017estimating, curth2021inductive}. In IHDP, the true propensities are unknown so we use estimated ones. Appendix C provides detailed description of both datasets for completeness. 

\textbf{Baselines.} We compare against state-of-the-art methods based on weighted conformal prediction (WCP) proposed in \cite{lei2021conformal}, which apply weighted CP to construct intervals for the two POs and combine them into intervals for ITEs. We consider the three variants of WCP in \cite{lei2021conformal}: (1) \textit{Na\"{i}ve WCP}, which combines the PO intervals using a Bonferroni correction, (2) \textit{Exact Nested WCP}, which uses WCP on one fold of the data to construct ITE intervals for the units in the other fold, by contrasting each unit's observed outcome with an interval for its missing PO, and then calibrates these intervals with a second conformal step, and (3) \textit{Inexact Nested WCP}, which follows the same steps as the exact version but replaces the second conformal step with conditional quantile regression. (Note that Inexact Nested WCP does not provide coverage guarantees.) For all methods, we use gradient boosting for nuisance and outcome modeling and the conformal quantile regression method in \cite{romano2019conformalized} to construct predictive intervals; the second stage of the nested WCP methods uses the quantile random forests of the implementation in \cite{lei2021conformal}. All methods use the true propensity score in the synthetic experiments, and a logistic regression estimate in IHDP and NLSM.

\vspace{-.1in}
\subsection{Results and discussion}
\label{Sec52}

\vspace{-.05in}
Our experimental findings yield the following key takeaways. Firstly, {\it the conformity scores of the IPW- and DR-learners dominate the oracle scores in the FOSD sense (i.e., $\alpha^*=1$) in all settings of Figures 4 and 5, which is stronger than the MCX guarantee in Theorem 2}, and both learners achieve the target coverage throughout; for the DR-learner, the two CDFs are close, most notably in IHDP. In contrast, the conformity scores of the X-learner are dominated by the oracle scores, and it under-covers in all experiments. Secondly, {\it the DR-learner provides the shortest intervals among the methods with coverage guarantees in most experiments} (all except NLSM); however, its point estimates are less accurate than those of the WCP baselines in Setup B, IHDP and NLSM, and on par with them in Setup A. Thirdly, {\it the effectiveness of conformal meta-learners depends on the discrepancy between the CDFs of conformity scores and oracle scores}: pseudo-outcome transformations that induce thicker tails in the conformity scores lead to overly conservative intervals (e.g., the IPW-learner, and the DR-learner in NLSM), whereas conformity scores that are stochastically smaller than the oracle scores lead to under-coverage (e.g., the X-learner).

{\bf Empirical assessment of stochastic orders.} Figure 4(a) depicts the empirical CDFs of the conformity scores $V_\varphi$ and oracle scores $V^*$ for the three meta-learners under study (DR-, IPW- and X-learners). These CDFs are averaged over 100 runs of {\bf Setups A} and {\bf B} of the synthetic generation process outlined in Section \ref{Sec51}. (The shaded regions represent the lowest and highest bounds on the empirical CDFs evaluated across all runs.) In both setups, the conformity scores of the DR- and IPW-learners demonstrate FOSD over the oracle scores with respect to the average CDFs. For the IPW-learner, FOSD also holds in almost every individual run; for the DR-learner, the CDFs of $V_\varphi$ and $V^*$ are close, and individual runs show small crossings (below 0.05 in 93\% of runs) that are consistent with sampling variability. This aligns with the result of Theorem 2, and shows that the stochastic dominance achieved in practice is even stronger than our theoretical guarantee, since FOSD (\mbox{\small $V_\varphi \succeq_{\mbox{\tiny (1)}} V^*$}) implies the weaker condition of convex dominance \mbox{\small $V_\varphi \succeq_{mcx} V^*$}. In contrast, the conformity scores of the X-learner are dominated by the oracle scores in the FOSD sense. This is not surprising in light of Theorem 2, which shows that the X-learner does not guarantee a distribution-free stochastic order. These observations are replicated in the semi-synthetic datasets (Figure 5), where the DR-learner's conformity scores are again close to the oracle scores in IHDP, and the X-learner's scores are only slightly dominated by the oracle scores in NLSM.

Based on Theorem 1, the FOSD observed in Figures 4(a) and 5 predicts that the IPW- and DR-learners will cover ITEs. Conversely, when the oracle scores dominate the conformity scores in the FOSD sense, applying the argument in the proof of Theorem 1(i) with the roles of $V_\varphi$ and $V^*$ reversed shows that coverage cannot exceed the target level by more than $1/(n_c+1)$, so the X-learner is not expected to over-cover. These predictions are confirmed by the results in Figures 4(b), 4(c) and Table 3: the IPW- and DR-learners achieve the target coverage at every level in Figure 4(c), whereas the X-learner under-covers at every level. The fact that the IPW- and DR-learners satisfy FOSD in these experiments is promising, because it indicates that their range of validity spans all levels of coverage (\mbox{\small $\alpha^*=1$} in Figure 3). However, FOSD cannot be guaranteed in a distribution-free manner.

\begin{figure}[t]
\arrayrulecolor{black}
\begin{minipage}{.6\textwidth}
\centering
\setlength{\tabcolsep}{3.5pt} 
\renewcommand{\arraystretch}{1} 
{\tiny 
\begin{tabular}{@{}rllllll@{}}
\toprule
\multicolumn{1}{c}{}       & \multicolumn{3}{c}{\textbf{IHDP}}                                                            & \multicolumn{3}{c}{\textbf{NLSM}}                                                            \\ \midrule
\multicolumn{1}{c}{}       & \multicolumn{1}{c}{{\bf Coverage}} & \multicolumn{1}{c}{{\bf Avg. len.}} & \multicolumn{1}{c}{{\bf RMSE}} & \multicolumn{1}{c}{{\bf Coverage}} & \multicolumn{1}{c}{{\bf Avg. len.}} & \multicolumn{1}{c}{{\bf RMSE}} \\ [1ex]
Na\"ive & 0.98 {\tiny (0.01)} & 8.88 {\tiny (0.34)} & {\bf 0.63} {\tiny (0.03)} & 1.00 {\tiny (0.00)} & 4.18 {\tiny (0.01)} & 0.13 {\tiny (0.00)} \\
Exact & 1.00 {\tiny (0.00)} & 10.0 {\tiny (0.24)} & 0.69 {\tiny (0.04)} & 1.00 {\tiny (0.00)} & 4.27 {\tiny (0.02)} & 0.15 {\tiny (0.00)} \\
Inexact & {\color{gray}{0.86 {\tiny (0.02)}}} & 5.08 {\tiny (0.23)} & 0.67 {\tiny (0.04)} & 0.93 {\tiny (0.00)} & {\bf 2.09} {\tiny (0.01)} & 0.15 {\tiny (0.00)} \\ \midrule 
X & {\color{gray}{0.83 {\tiny (0.01)}}} & 4.44 {\tiny (0.08)} & 0.77 {\tiny (0.05)} & {\color{gray}{0.89 {\tiny (0.00)}}} & 1.76 {\tiny (0.01)} & {\bf 0.12} {\tiny (0.00)} \\
IPW & 1.00 {\tiny (0.00)} & 29.2 {\tiny (1.99)} & 5.18 {\tiny (0.49)} & 1.00 {\tiny (0.00)} & 6.34 {\tiny (0.06)} & 0.72 {\tiny (0.02)} \\
DR & 0.97 {\tiny (0.01)} & {\bf 7.98} {\tiny (0.21)} & 1.08 {\tiny (0.07)} & 1.00 {\tiny (0.00)} & 5.21 {\tiny (0.05)} & 0.26 {\tiny (0.01)}
 \\ \bottomrule \vspace{.005in}
\end{tabular}}

{\footnotesize Table 3: Performance of all baselines in semi-synthetic datasets. Bold numbers correspond to best performance.}
\end{minipage}
\begin{minipage}{.4\textwidth}
\centering
\includegraphics[width=2.2in]{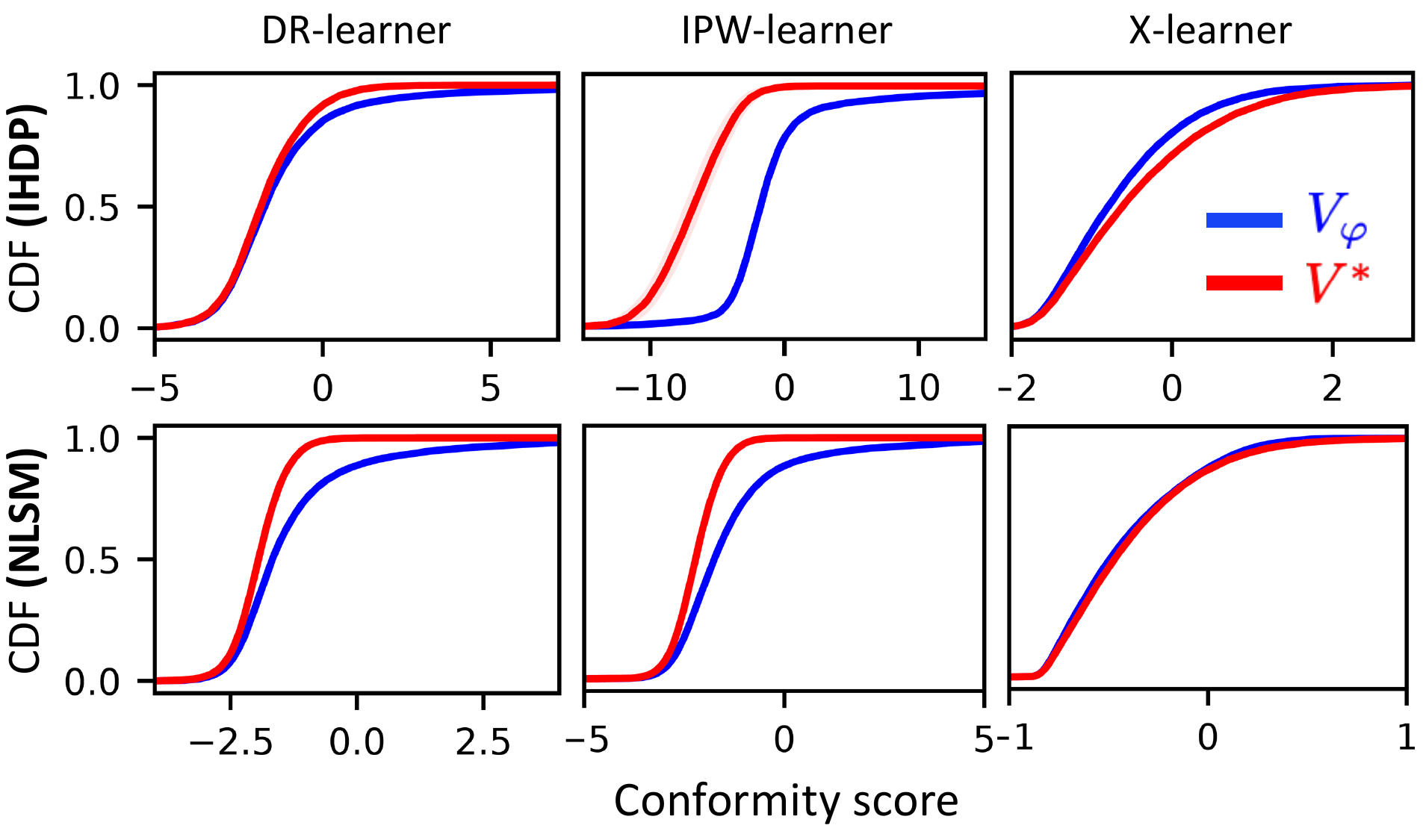}

{\footnotesize Figure 5: Stochastic orders in IHDP/NLSM.}
\end{minipage}
\vspace{-.225in} 
\end{figure}

{\bf Coverage, efficiency and point estimation accuracy.} The performance of a predictive inference procedure can be characterized in terms of three metrics: achieved coverage for true ITEs, expected length of predictive intervals, and root-mean-square error (RMSE) in CATE estimates. The DR-learner achieves the target coverage in all experiments and provides the shortest intervals among the methods with coverage guarantees in most of them, whereas the WCP baselines generally provide more accurate point estimates (see Appendix C for further experiments). In Figure 4(b), we can see that the DR-learner achieves the target coverage of 90\% and provides shorter intervals than the valid (na\"ive and exact) WCP procedures in both setups. Its RMSE is on par with exact WCP in {\bf Setup A}, but higher than that of both valid WCP procedures in {\bf Setup B}. The X-learner achieves the lowest RMSE in {\bf Setup A}, but, as expected, it under-covers ITEs in all experiments. The inexact WCP baseline offers competitive efficiency and RMSE in both setups, but it does not offer coverage guarantees and empirically it falls short of the target coverage in {\bf Setup B}. Figure 4(c) shows that the efficiency advantage of the DR-learner over the valid WCP procedures holds at all levels of target coverage, whereas its RMSE relative to WCP depends on the setup, i.e., it is lower at most coverage levels in {\bf Setup A}, and higher at high coverage levels in {\bf Setup B}.

The semi-synthetic experiments on IHDP and NLSM datasets shed light on when meta-learners may perform poorly. On IHDP, the DR-learner achieves the desired coverage of 90\% with the shortest intervals among the methods that reach this coverage, but its RMSE is higher than that of the WCP baselines and the X-learner (Table 3). We observe that interval efficiency depends on how closely the CDF of conformity scores matches the oracle CDF. The DR-learner produces longer intervals than the valid WCP procedures when its conformity scores have a strong dominance over the oracle scores, as observed in the NLSM dataset (Figure 5, bottom). Conversely, when the CDF of conformity scores is a closer lower bound on the oracle CDF, the DR-learner's intervals are the shortest among the valid methods (Figure 4 and Figure 5, top). This is intuitive because if the pseudo-outcome transformation induces significant variability in regression targets, it will result in a lower CDF, poorer accuracy of pseudo-outcome regression, and longer predictive intervals. This is why the DR-learner provides shorter intervals than the IPW-learner in all experiments, and lower RMSE in all except {\bf Setup B}, i.e., the DR-learner provides a better approximation of the oracle CDF. Future work could focus on analyzing the gap between pseudo-outcome and oracle score CDFs and designing pseudo-outcome transformations that optimize efficiency while preserving stochastic orders.

\section{Conclusions}
\vspace{-.05in} 

Estimation and inference of treatment effects is challenging because causal effects are not directly observable. In this paper, we developed a general framework for inference of treatment effects, dubbed conformal meta-learners, that is compatible with any machine learning model. Our framework inherits the model- and distribution-free properties of conformal prediction, as well as the estimation accuracy of model-agnostic meta-learners of treatment effects. Additionally, we introduce a new theoretical framework based on stochastic ordering to assess the validity of our method, which can guide the development of new models optimized for both accurate estimation and valid inference.

\bibliographystyle{unsrt}
\bibliography{conformalITE}

@article{rubin1990application,
  title={[On the application of probability theory to agricultural experiments. Essay on principles. Section 9.] Comment: Neyman (1923) and causal inference in experiments and observational studies},
  author={Rubin, Donald B},
  journal={Statistical Science},
  volume={5},
  number={4},
  pages={472--480},
  year={1990},
  publisher={JSTOR}
}

@article{alaa2017bayesian,
  title={Bayesian inference of individualized treatment effects using multi-task gaussian processes},
  author={Alaa, Ahmed M and Van Der Schaar, Mihaela},
  journal={Advances in neural information processing systems},
  volume={30},
  year={2017}
}

@article{yeager2019national,
  title={A national experiment reveals where a growth mindset improves achievement},
  author={Yeager, David S and Hanselman, Paul and Walton, Gregory M and Murray, Jared S and Crosnoe, Robert and Muller, Chandra and Tipton, Elizabeth and Schneider, Barbara and Hulleman, Chris S and Hinojosa, Cintia P and others},
  journal={Nature},
  volume={573},
  number={7774},
  pages={364--369},
  year={2019},
  publisher={Nature Publishing Group}
}

@article{greenfield2007heterogeneity,
  title={Heterogeneity of treatment effects: implications for guidelines, payment, and quality assessment},
  author={Greenfield, Sheldon and Kravitz, Richard and Duan, Naihua and Kaplan, Sherrie H},
  journal={The American journal of medicine},
  volume={120},
  number={4},
  pages={S3--S9},
  year={2007},
  publisher={Elsevier}
}

@article{imai2011estimation,
  title={Estimation of heterogeneous treatment effects from randomized experiments, with application to the optimal planning of the get-out-the-vote campaign},
  author={Imai, Kosuke and Strauss, Aaron},
  journal={Political Analysis},
  volume={19},
  number={1},
  pages={1--19},
  year={2011},
  publisher={Cambridge University Press}
}

@article{neyman1923application,
  title={On the application of probability theory to agricultural experiments. essay on principles. section 9.(tlanslated and edited by dm dabrowska and tp speed, statistical science (1990), 5, 465-480)},
  author={Neyman, Jerzy S},
  journal={Annals of Agricultural Sciences},
  volume={10},
  pages={1--51},
  year={1923}
}

@article{kunzel2019metalearners,
  title={Metalearners for estimating heterogeneous treatment effects using machine learning},
  author={K{\"u}nzel, S{\"o}ren R and Sekhon, Jasjeet S and Bickel, Peter J and Yu, Bin},
  journal={Proceedings of the national academy of sciences},
  volume={116},
  number={10},
  pages={4156--4165},
  year={2019},
  publisher={National Acad Sciences}
}

@article{horvitz1952generalization,
  title={A generalization of sampling without replacement from a finite universe},
  author={Horvitz, Daniel G and Thompson, Donovan J},
  journal={Journal of the American statistical Association},
  volume={47},
  number={260},
  pages={663--685},
  year={1952},
  publisher={Taylor \& Francis}
}

@article{kennedy2020towards,
  title={Towards optimal doubly robust estimation of heterogeneous causal effects},
  author={Kennedy, Edward H},
  journal={arXiv preprint arXiv:2004.14497},
  year={2020}
}

@inproceedings{johansson2016learning,
  title={Learning representations for counterfactual inference},
  author={Johansson, Fredrik and Shalit, Uri and Sontag, David},
  booktitle={International conference on machine learning},
  pages={3020--3029},
  year={2016},
  organization={PMLR}
}

@inproceedings{shalit2017estimating,
  title={Estimating individual treatment effect: generalization bounds and algorithms},
  author={Shalit, Uri and Johansson, Fredrik D and Sontag, David},
  booktitle={International Conference on Machine Learning},
  pages={3076--3085},
  year={2017},
  organization={PMLR}
}

@article{wager2018estimation,
  title={Estimation and inference of heterogeneous treatment effects using random forests},
  author={Wager, Stefan and Athey, Susan},
  journal={Journal of the American Statistical Association},
  volume={113},
  number={523},
  pages={1228--1242},
  year={2018},
  publisher={Taylor \& Francis}
}

@article{hill2011bayesian,
  title={Bayesian nonparametric modeling for causal inference},
  author={Hill, Jennifer L},
  journal={Journal of Computational and Graphical Statistics},
  volume={20},
  number={1},
  pages={217--240},
  year={2011},
  publisher={Taylor \& Francis}
}

@article{carvalho2019assessing,
  title={Assessing treatment effect variation in observational studies: Results from a data challenge},
  author={Carvalho, Carlos and Feller, Avi and Murray, Jared and Woody, Spencer and Yeager, David},
  journal={Observational Studies},
  volume={5},
  number={2},
  pages={21--35},
  year={2019},
  publisher={University of Pennsylvania Press}
}

@book{shaked2007stochastic,
  title={Stochastic orders},
  author={Shaked, Moshe and Shanthikumar, J George},
  year={2007},
  publisher={Springer}
}

@article{hadar1969rules,
  title={Rules for ordering uncertain prospects},
  author={Hadar, Josef and Russell, William R},
  journal={The American economic review},
  volume={59},
  number={1},
  pages={25--34},
  year={1969},
  publisher={JSTOR}
}

@incollection{rothschild1978increasing,
  title={Increasing risk: I. A definition},
  author={Rothschild, Michael and Stiglitz, Joseph E},
  booktitle={Uncertainty in Economics},
  pages={99--121},
  year={1978},
  publisher={Elsevier}
}

@book{asmussen2020risk,
  title={Risk and insurance},
  author={Asmussen, S{\o}ren and Steffensen, Mogens},
  year={2020},
  publisher={Springer}
}

@article{brooks1992effects,
  title={Effects of early intervention on cognitive function of low birth weight preterm infants},
  author={Brooks-Gunn, Jeanne and Liaw, Fong-ruey and Klebanov, Pamela Kato},
  journal={The Journal of pediatrics},
  volume={120},
  number={3},
  pages={350--359},
  year={1992},
  publisher={Elsevier}
}

@inproceedings{linusson2014signed,
  title={Signed-error conformal regression},
  author={Linusson, Henrik and Johansson, Ulf and L{\"o}fstr{\"o}m, Tuve},
  booktitle={Advances in Knowledge Discovery and Data Mining: 18th Pacific-Asia Conference, PAKDD 2014, Tainan, Taiwan, May 13-16, 2014. Proceedings, Part I 18},
  pages={224--236},
  year={2014},
  organization={Springer}
}

@article{lei2021conformal,
  title={Conformal inference of counterfactuals and individual treatment effects},
  author={Lei, Lihua and Cand{\`e}s, Emmanuel J},
  journal={Journal of the Royal Statistical Society Series B: Statistical Methodology},
  volume={83},
  number={5},
  pages={911--938},
  year={2021},
  publisher={Oxford University Press}
}

@article{tibshirani2019conformal,
  title={Conformal prediction under covariate shift},
  author={Tibshirani, Ryan J and Barber, Rina Foygel and Cand{\`e}s, Emmanuel J and Ramdas, Aaditya},
  journal={arXiv preprint arXiv:1904.06019},
  year={2019}
}

@article{vovk1999machine,
  title={Machine-learning applications of algorithmic randomness},
  author={Vovk, Volodya and Gammerman, Alexander and Saunders, Craig},
  journal={ International Conference on Machine Learning},
  year={1999},
  publisher={Citeseer}
}

@article{lei2013distribution,
  title={Distribution-free prediction sets},
  author={Lei, Jing and Robins, James and Wasserman, Larry},
  journal={Journal of the American Statistical Association},
  volume={108},
  number={501},
  pages={278--287},
  year={2013},
  publisher={Taylor \& Francis}
}

@article{lei2018distribution,
  title={Distribution-free predictive inference for regression},
  author={Lei, Jing and G’Sell, Max and Rinaldo, Alessandro and Tibshirani, Ryan J and Wasserman, Larry},
  journal={Journal of the American Statistical Association},
  volume={113},
  number={523},
  pages={1094--1111},
  year={2018},
  publisher={Taylor \& Francis}
}

@book{papadopoulos2008inductive,
  title={Inductive conformal prediction: Theory and application to neural networks},
  author={Papadopoulos, Harris},
  year={2008},
  publisher={INTECH Open Access Publisher Rijeka}
}

@article{sniekers2015adaptive,
  title={Adaptive Bayesian credible sets in regression with a Gaussian process prior},
  author={Sniekers, Suzanne and van der Vaart, Aad},
  journal={Electronic Journal of Statistics},
  volume={9},
  number={2},
  pages={2475--2527},
  year={2015},
  publisher={Institute of Mathematical Statistics and Bernoulli Society}
}

@article{gibbs2021adaptive,
  title={Adaptive conformal inference under distribution shift},
  author={Gibbs, Isaac and Candes, Emmanuel},
  journal={Advances in Neural Information Processing Systems},
  volume={34},
  pages={1660--1672},
  year={2021}
}

@article{jin2023sensitivity,
  title={Sensitivity analysis of individual treatment effects: A robust conformal inference approach},
  author={Jin, Ying and Ren, Zhimei and Cand{\`e}s, Emmanuel J},
  journal={Proceedings of the National Academy of Sciences},
  volume={120},
  number={6},
  pages={e2214889120},
  year={2023},
  publisher={National Acad Sciences}
}

@inproceedings{makar2020estimation,
  title={Estimation of bounds on potential outcomes for decision making},
  author={Makar, Maggie and Johansson, Fredrik and Guttag, John and Sontag, David},
  booktitle={International Conference on Machine Learning},
  pages={6661--6671},
  year={2020},
  organization={PMLR}
}

@article{taufiq2022conformal,
  title={Conformal Off-Policy Prediction in Contextual Bandits},
  author={Taufiq, Muhammad Faaiz and Ton, Jean-Francois and Cornish, Rob and Teh, Yee Whye and Doucet, Arnaud},
  journal={arXiv preprint arXiv:2206.04405},
  year={2022}
}

@inproceedings{zhang2023conformal,
  title={Conformal Off-Policy Prediction},
  author={Zhang, Yingying and Shi, Chengchun and Luo, Shikai},
  booktitle={International Conference on Artificial Intelligence and Statistics},
  pages={2751--2768},
  year={2023},
  organization={PMLR}
}

@article{qiu2022distribution,
  title={Distribution-free prediction sets adaptive to unknown covariate shift},
  author={Qiu, Hongxiang and Dobriban, Edgar and Tchetgen, Eric Tchetgen},
  journal={arXiv preprint arXiv:2203.06126},
  year={2022}
}

@article{yin2022conformal,
  title={Conformal sensitivity analysis for individual treatment effects},
  author={Yin, Mingzhang and Shi, Claudia and Wang, Yixin and Blei, David M},
  journal={Journal of the American Statistical Association},
  pages={1--14},
  year={2022},
  publisher={Taylor \& Francis}
}

@article{barber2022conformal,
  title={Conformal prediction beyond exchangeability},
  author={Barber, Rina Foygel and Candes, Emmanuel J and Ramdas, Aaditya and Tibshirani, Ryan J},
  journal={arXiv preprint arXiv:2202.13415},
  year={2022}
}

@article{wager2014confidence,
  title={Confidence intervals for random forests: The jackknife and the infinitesimal jackknife},
  author={Wager, Stefan and Hastie, Trevor and Efron, Bradley},
  journal={The Journal of Machine Learning Research},
  volume={15},
  number={1},
  pages={1625--1651},
  year={2014},
  publisher={JMLR. org}
}

@article{curth2021inductive,
  title={On inductive biases for heterogeneous treatment effect estimation},
  author={Curth, Alicia and van der Schaar, Mihaela},
  journal={Advances in Neural Information Processing Systems},
  volume={34},
  pages={15883--15894},
  year={2021}
}

@article{szabo2015frequentist,
  title={Frequentist coverage of adaptive nonparametric Bayesian credible sets},
  author={Szab{\'o}, Botond and Van Der Vaart, Aad W and Van Zanten, JH},
  journal={The Annals of Statistics},
  year={2015}
}

@inproceedings{ganin2015unsupervised,
  title={Unsupervised domain adaptation by backpropagation},
  author={Ganin, Yaroslav and Lempitsky, Victor},
  booktitle={International conference on machine learning},
  pages={1180--1189},
  year={2015},
  organization={PMLR}
}

@article{shimodaira2000improving,
  title={Improving predictive inference under covariate shift by weighting the log-likelihood function},
  author={Shimodaira, Hidetoshi},
  journal={Journal of statistical planning and inference},
  volume={90},
  number={2},
  pages={227--244},
  year={2000},
  publisher={Elsevier}
}

@book{imbens2015causal,
  title={Causal inference in statistics, social, and biomedical sciences},
  author={Imbens, Guido W and Rubin, Donald B},
  year={2015},
  publisher={Cambridge University Press}
}

@article{li2019visualbert,
  title={Visualbert: A simple and performant baseline for vision and language},
  author={Li, Liunian Harold and Yatskar, Mark and Yin, Da and Hsieh, Cho-Jui and Chang, Kai-Wei},
  journal={arXiv preprint arXiv:1908.03557},
  year={2019}
}

@article{chipman2010bart,
  title={BART: Bayesian additive regression trees},
  author={Chipman, Hugh A and George, Edward I and McCulloch, Robert E},
  journal={The Annals of Applied Statistics},
  volume={4},
  number={1},
  pages={266--298},
  year={2010},
  publisher={Institute of Mathematical Statistics}
}

@article{hahn2020bayesian,
  title={Bayesian regression tree models for causal inference: Regularization, confounding, and heterogeneous effects (with discussion)},
  author={Hahn, P Richard and Murray, Jared S and Carvalho, Carlos M},
  journal={Bayesian Analysis},
  volume={15},
  number={3},
  pages={965--1056},
  year={2020},
  publisher={International Society for Bayesian Analysis}
}

@article{romano2019conformalized,
  title={Conformalized quantile regression},
  author={Romano, Yaniv and Patterson, Evan and Candes, Emmanuel},
  journal={Advances in Neural Information Processing Systems},
  volume={32},
  pages={3543--3553},
  year={2019}
}

@book{vovk2005algorithmic,
  title={Algorithmic learning in a random world},
  author={Vovk, Vladimir and Gammerman, Alexander and Shafer, Glenn},
  year={2005},
  publisher={Springer Science \& Business Media}
}

\newpage
\appendix

\renewcommand{\theequation}{A.\arabic{equation}}
\setcounter{equation}{0}

\section*{Appendix A: Technical Proofs}

\subsection*{A.1. Equivalent Definitions for Stochastic Orders}

The following section introduces alternative definitions for stochastic dominance between two cumulative distribution functions (CDFs), denoted as $F$ and $G$. These definitions will be used in the proof for Theorem 2 (Section A.4.). Note that these definitions are equivalent to Definition 1 provided in the main text but are expressed in decision-theoretic terms.

{\bf Definition A1} \textbf{(First-order stochastic dominance)} $F$ has {\it first-order} stochastic dominance (FOSD) over $G$, $F \succeq_{\mbox{\tiny (1)}}G$, if and only if $\mathbb{E}_{X\sim F}[u(X)] \geq \mathbb{E}_{X\sim G}[u(X)]$ for all non-decreasing~(utility)~functions $u: \mathbb{R}\to\mathbb{R}$ (Theorem 1.A.3 in \cite{shaked2007stochastic} proves the equivalence of Definitions 1 and A1).

{\bf Definition A2} \textbf{(Second-order stochastic dominance)} $F$ has {\it second-order} stochastic dominance (SOSD) over $G$, $F \succeq_{\mbox{\tiny (2)}}G$, if and only if $\mathbb{E}_{X\sim F}[u(X)] \geq \mathbb{E}_{X\sim G}[u(X)]$ for~all~non-decreasing~concave functions $u: \mathbb{R}\to\mathbb{R}$ (Refer to Theorem 4.A.1 and Eq. (4.A.7) in \cite{shaked2007stochastic}).

In addition to the alternative definitions of FOSD and SOSD above, we also state the definitions for two additional notions of stochastic ordering that we will use in the proofs. 

{\bf Definition A3} \textbf{(Statewise stochastic dominance)} A random variable~$X$~has~{\it statewise}~stochastic~dominance (SWD) over a random variable $Y$, $X \succeq_{\mbox{\tiny SWD}}Y$, if and only if $X \geq Y$ for every draw of the bivariate random variable $(X, Y) \sim P_{X, Y}$. Note that SWD implies FOSD.

{\bf Definition A4} \textbf{(Convex dominance)} $F$ has {\it convex} dominance (CX) over $G$, $F \succeq_{\mbox{\tiny cx}}G$, if and only if $\mathbb{E}_{X\sim F}[u(X)] \geq \mathbb{E}_{X\sim G}[u(X)]$ for~all convex functions $u: \mathbb{R}\to\mathbb{R}$.

\subsection*{A.2. Useful Lemmas}

{\bf Lemma A1. (Pointwise and marginal stochastic orders)} {\it For random variables $X$, $Y$, and $Z$, the following conditions hold for any marginal distribution $P_X$: 
\begin{enumerate}[(i)]
    \item {\it $Y\,|\,X=x\,\, \succeq_{\mbox{\tiny (1)}}Z\,|\,X=x, \forall x \in \mathcal{X}\,\,\,\,\,\,  \Rightarrow\,\, Y\, \succeq_{\mbox{\tiny (1)}} Z,$}
    \item {\it $Y\,|\,X=x\,\, \succeq_{\mbox{\tiny (2)}} Z\,|\,X=x, \forall x \in \mathcal{X}\,\,\,\,\,\, \Rightarrow\,\, Y\, \succeq_{\mbox{\tiny (2)}} Z,$}
    \item {\it $Y\,|\,X=x\,\, \succeq_{mcx}\, Z\,|\,X=x, \forall x \in \mathcal{X} \Rightarrow\,\, Y\, \succeq_{mcx} Z.$}
\end{enumerate}}

{\it Proof.}\,\, Recall from the definition of FOSD (Definition A1) that if $Y\,|\,X=x \succeq_{\mbox{\tiny (1)}}Z\,|\,X=x$, then:
\begin{equation}
    \mathbb{E}_{V\sim P_{Y|X=x}}[u(V)] \geq \mathbb{E}_{W\sim P_{Z|X=x}}[u(W)],\, \forall x \in \mathcal{X},
    \label{eqA1}
\end{equation}
for all non-decreasing functions $u: \mathbb{R}\to\mathbb{R}$. Since  expectation is a positive~linear~operator,~marginalizing both sides with respect to the density $P_X$ preserves the inequality in (\ref{eqA1}):
\begin{equation}
    \mathbb{E}_{X \sim P_X}[\mathbb{E}_{V\sim P_{Y|X}}[u(V)]] \geq \mathbb{E}_{X \sim P_X}[\mathbb{E}_{W\sim P_{Z|X}}[u(W)]],
    \label{eqA2}
\end{equation}
which recovers the definition of FOSD between random variables $Y$ and $Z$, i.e., 
\begin{equation}
    \mathbb{E}_{Y\sim P_{Y}}[u(Y)] \geq \mathbb{E}_{Z\sim P_{Z}}[u(Z)],
    \label{eqA3}
\end{equation}
and concludes the proof for (i). Note that (\ref{eqA3}) holds for all non-decreasing functions $u: \mathbb{R}\to\mathbb{R}$, which also includes all non-decreasing concave and convex functions that define SOSD and MCX. Hence, combining (\ref{eqA1})-(\ref{eqA3}) with Definitions 2 and A2 concludes the proof for (ii) and (iii). \qed

\vspace{.1in}
{\bf Lemma A2. (Convex dominance of mixture distributions over convex combinations of random variables)} {\it Let $X \in \mathbb{R}$ and $Y \in \mathbb{R}$ be two real-valued random variables with finite means, and let $\pi \in [0, 1]$ be a constant. If $Z$ and $V$ are two random variables constructed as:
\begin{itemize}
    \item $Z$ is the absolute value of the convex combination of $X$ and $Y$, i.e., $Z = |\pi X + (1-\pi) Y|$,
    \item $V$ is a mixture of $|X|$ and $|Y|$, i.e., $V = W |X| + (1-W)|Y|,\, W \sim \mbox{Bernoulli}(\pi)$ independent of $(X,Y)$,
\end{itemize}
}
then we have $V \succeq_{mcx} Z$ for any $\pi \in [0, 1]$ and any distribution $P(X, Y)$.

{\it Proof.}\,\, Recall from Definition 2 that $V \succeq_{mcx} Z$ if and only if
\begin{equation}
\mathbb{E}_{V}[u(V)] \geq \mathbb{E}_{Z}[u(Z)], 
\end{equation}
for all non-decreasing convex functions $u: \mathbb{R}\to\mathbb{R}$. Given the definition of $V$ above, we have
\begin{align}
\mathbb{E}_{V}[u(V)] &= P(W=1)\cdot\mathbb{E}_V[u(V)\,|\,W=1] + P(W=0)\cdot\mathbb{E}_V[u(V)\,|\,W=0] \nonumber \\
&= \pi\cdot\mathbb{E}[u(|X|)] + (1-\pi)\cdot\mathbb{E}[u(|Y|)] \nonumber \\
&\overset{(a)}{\geq} \mathbb{E}[u(\pi |X| + (1-\pi) |Y|)] = \mathbb{E}[u(|\pi X| + |(1-\pi) Y|)] \nonumber \\
&\overset{(b)}{\geq} \mathbb{E}[u(|\pi X + (1-\pi) Y|)] = \mathbb{E}[u(Z)] \, \Rightarrow V \succeq_{mcx} Z,
\end{align}
where the last two inequalities follow from (a) the convexity of $u$, and (b) the monotonicity~of~$u$~along with the application of the triangle inequality $|\pi X| + |(1-\pi) Y| \geq |\pi X + (1-\pi) Y|$. \qed

\vspace{.1in}
{\bf Lemma A3. (Stochastic dominance and sign changes in CDFs)} {\it Let $X$ and $Y$ be two non-negative random variables with distribution functions $F$ and $G$ and with
finite means such that $\mathbb{E}[X] \leq \mathbb{E}[Y]$. Let $S^-(f)$ be the number of sign changes of a function $f$ defined as}
\begin{align}
S^-(f) = \sup S^-[f(x_1), f(x_2), \ldots, f(x_m)],
\label{eqA6}
\end{align}
{\it where $S^-[a_1, a_2, \ldots, a_m]$ is the number of sign changes of the sequence $[a_1, a_2, \ldots, a_m]$, with the zero terms being discarded, and the supremum in (\ref{eqA6}) is taken over all sets $x_1 < x_2 < \ldots < x_m$ such that $m < \infty$. Then, $F \succeq_{\mbox{\tiny (2)}}[\succeq_{mcx}] G$ if and only if there exist random variables $Z_1$, $Z_2$,\ldots, with distribution functions $G_1$, $G_2$,\ldots, such that $Z_1 =_{st} X$, $\mathbb{E}[Z_j] \leq \mathbb{E}[Y]$, $j=1,2,\ldots$, $Z_j \to_{st} Y$ and $\mathbb{E}[Z_j] \to \mathbb{E}[Y]$ as $j \to \infty$ and $S^-(G_{j+1} - G_j)=1$ and the sign sequence is $\{+,-\} [\{-,+\}]$. Here, $=_{st}$ denotes equality in law and $\to_{st}$ denotes convergence in distribution.}

{\it Proof.}\,\, Refer to Theorems 4.A.22 and 4.A.23 in \cite{shaked2007stochastic} for a full proof.  \qed 

\vspace{.1in}
{\bf Lemma A4. (CDF crossings under MCX)} {\it Let $X$ and $Y$ be random variables with CDFs $F$ and $G$ and finite means, such that $F \succeq_{mcx} G$ and $S^-(F-G)<\infty$. Let $v^* \triangleq \inf\{v: F(u)\leq G(u),\, \forall u \geq v\}$. Then $v^*<\infty$, $F(v)\leq G(v)$ for all $v \geq v^*$, and $F(v^*)<1$ (with the convention $F(-\infty)=0$).}

{\it Proof.}\,\, Applying Definition 2 to the non-decreasing convex functions $u(x)=(x-t)_+$, $t \in \mathbb{R}$, gives
\begin{equation}
\int_t^\infty \left[G(v)-F(v)\right] dv = \mathbb{E}[(X-t)_+]-\mathbb{E}[(Y-t)_+] \geq 0,\,\, \forall t \in \mathbb{R}.
\tag{$\star$}
\label{eqA4star}
\end{equation}
Since $S^-(F-G)<\infty$, there exists $w$ such that the nonzero values of $F-G$ on $[w,\infty)$ all have the same sign. If this sign were positive, there would be $t \geq w$ with $F(t)>G(t)$ and, by right-continuity of CDFs, $F>G$ on an interval $[t, t+\eta)$, $\eta>0$; the integrand in (\ref{eqA4star}) would then be non-positive on $[t,\infty)$ and negative on $[t,t+\eta)$, contradicting (\ref{eqA4star}). Hence, $F\leq G$ on $[w,\infty)$, so that $v^*\leq w<\infty$ and, by right-continuity, $F(v)\leq G(v)$ for all $v\geq v^*$. Now suppose that $v^*>-\infty$ and $F(v^*)=1$. Then $F=G=1$ on $[v^*,\infty)$ and, by definition of $v^*$, every interval $[v, v^*)$ contains a point $t$ with $F(t)>G(t)$. Since $S^-(F-G)<\infty$, it follows that $F-G \geq 0$ on $[v^*-\delta, v^*)$ for some $\delta>0$. Taking $t\in[v^*-\delta,v^*)$ with $F(t)>G(t)$ yields $\int_t^\infty [G(v)-F(v)]\,dv = \int_t^{v^*} [G(v)-F(v)]\,dv<0$, contradicting (\ref{eqA4star}). Hence, $F(v^*)<1$. \qed

\subsection*{A.3. Proof of Theorem 1} 

{\bf Theorem 1.} {\it If $(X_i, W_i, Y_i(0), Y_i(1)),\, i=1,\ldots, n+1$ are i.i.d., then $\exists\, \alpha^* \in (0, 1]$ such that the pseudo-interval $\widehat{C}_\varphi(X_{n+1})$ constructed using the dataset $\mathcal{D}=\{(X_i, W_i, Y_i)\}_{i=1}^n$ satisfies} 
\begin{align}
\mathbb{P}(Y_{n+1}(1)-Y_{n+1}(0)\in\widehat{C}_\varphi(X_{n+1})\,|\,\mathcal{D}_\varphi\cup\mathcal{D}_t) \geq 1-\alpha-\epsilon_{n_c}(\alpha),\, \forall \alpha \in (0, \alpha^*), \nonumber
\end{align}
{\it if at least one of the following stochastic ordering conditions hold for the distributions of $V_\varphi$ and $V^*$ conditional on $\mathcal{D}_\varphi\cup\mathcal{D}_t$: (i) $V_\varphi \succeq_{\mbox{\tiny (1)}} V^*$,~or (ii) $V_\varphi \succeq_{mcx} V^*$ and $F_{V_\varphi}-F_{V^*}$ changes sign finitely many times. Under (i), we have $\alpha^*=1$ and $\epsilon_{n_c}(\alpha)=0$.~Under~(ii), $\epsilon_{n_c}(\alpha)=\exp(-2n_c(\alpha^*-\alpha)^2)$, where $n_c=|\mathcal{D}_c|$ and $\alpha^*=1-F_{V_\varphi}(v^*)$, with $v^*$ the last crossing point of $F_{V_\varphi}$ and $F_{V^*}$.}

\vspace{.05in}
{\it Proof.}\,\, Throughout the proof, we condition on $\mathcal{D}_\varphi\cup\mathcal{D}_t$, so that $\widehat{\varphi}$ and $\widehat{\tau}$ are fixed and all probabilities are conditional on $\mathcal{D}_\varphi\cup\mathcal{D}_t$. Since the data are i.i.d., the test scores $(V_{\varphi,n+1}, V^*_{n+1})$ are then independent of the calibration scores, and $V_{\varphi,n+1}$ is exchangeable with the calibration scores. Without loss of generality, assume that the conformity scores are sorted, $V_{\varphi,1} < \ldots < V_{\varphi,n_c},$ where $n_c = |\mathcal{D}_c|$. Recall from (\ref{eq10b}) that the pseudo-interval is constructed as:
\renewcommand{\theequation}{A.\arabic{equation}}
\begin{equation}
\widehat{C}_\varphi(x) = [\,\widehat{\tau}(x) - Q_{\mathcal{V}_\varphi}(1-\alpha),\,\widehat{\tau}(x) + Q_{\mathcal{V}_\varphi}(1-\alpha)\,],
\label{eqA7}
\end{equation}
where $Q_{\mathcal{V}_\varphi}(1-\alpha)$ is the empirical quantile defined as
\begin{equation}
Q_{\mathcal{V}_\varphi}(1-\alpha) = 
\left\{
\begin{array}{ll}
      V_{\varphi,\lceil (n_c+1)(1-\alpha) \rceil}, & \alpha \geq \frac{1}{n_c+1}, \\
      \infty, & o.w. \\
\end{array} 
\right.
\label{eqA8}
\end{equation}
Combining (\ref{eqA7}) and (\ref{eqA8}), we notice that the following two events are equivalent
\begin{equation}
\{Y_{n+1}(1)-Y_{n+1}(0) \in \widehat{C}_\varphi(X_{n+1})\} \iff \{V^*_{n+1} \leq Q_{\mathcal{V}_\varphi}(1-\alpha)\},
\label{eqA9}
\end{equation}
or equivalently, 
\begin{equation}
\{Y_{n+1}(1)-Y_{n+1}(0) \in \widehat{C}_\varphi(X_{n+1})\} \iff \{V^*_{n+1} \leq V_{\varphi,\lceil (n_c+1)(1-\alpha) \rceil}\}.
\label{eqA10}
\end{equation}
Hence, the achieved coverage probability for the pseudo-interval is given by
\begin{equation}
\mathbb{P}(V^*_{n+1} \leq V_{\varphi,\lceil (n_c+1)(1-\alpha) \rceil}).
\label{eqA11}
\end{equation}
By exchangeability of the variables $(X_1, Y_1(1)-Y_1(0)), . . . ,(X_{n+1}, Y_{n+1}(1)-Y_{n+1}(0))$, we have
\begin{equation}
\mathbb{P}(V_{\varphi, n+1} \leq V_{\varphi,k}) = \frac{k}{n_c+1},
\label{eqA12}
\end{equation}
for any integer $k$. (\ref{eqA12}) holds because ranks are uniformly distributed under exchangeability, i.e., $V_{\varphi, n+1}$ is equally likely to fall in anywhere between the calibration points $V_{\varphi, 1},\ldots, V_{\varphi, n_c}$. From (\ref{eqA12}), it follows that for $\lceil (n_c+1)(1-\alpha) \rceil$ we have: 
\begin{equation}
\mathbb{P}(V_{\varphi, n+1} \leq V_{\varphi,\lceil (n_c+1)(1-\alpha) \rceil}) = \frac{\lceil (n_c+1)(1-\alpha) \rceil}{n_c+1} \geq 1-\alpha. 
\label{eqA13}
\end{equation}
Now, we examine the two conditions: (i) $V_\varphi \succeq_{\mbox{\tiny (1)}} V^*$, and (ii) $V_\varphi \succeq_{mcx} V^*$.\\

{\bf (i) FOSD $\boldsymbol{V_\varphi \succeq_{\mbox{\tiny (1)}} V^*}$:}

\vspace{.05in}
If $V_\varphi \succeq_{\mbox{\tiny (1)}} V^*$, then from Definition 1: $F_{V_\varphi}(v) \leq F_{V^*}(v), \forall v$. Equivalently, FOSD can be written as:
\begin{equation}
V_\varphi \succeq_{\mbox{\tiny (1)}} V^* \iff \mathbb{P}(V_\varphi \leq v) \leq \mathbb{P}(V^* \leq v),\, \forall v.
\label{eqA14}
\end{equation}
Since (\ref{eqA14}) applies for any $v$, and $V_{\varphi,n+1}$ and $V^*_{n+1}$ are independent of the calibration scores, then the following holds:
\begin{equation}
V_{\varphi, n+1} \succeq_{\mbox{\tiny (1)}} V^*_{n+1} \Rightarrow \mathbb{P}(V_{\varphi, n+1} \leq V_{\varphi,\lceil (n_c+1)(1-\alpha) \rceil}) \leq \mathbb{P}(V^*_{n+1} \leq V_{\varphi,\lceil (n_c+1)(1-\alpha) \rceil}). 
\label{eqA15}
\end{equation}
Combining (\ref{eqA13}) and (\ref{eqA15}) we have
\begin{equation}
V_{\varphi, n+1} \succeq_{\mbox{\tiny (1)}} V^*_{n+1} \Rightarrow \mathbb{P}(V^*_{n+1} \leq V_{\varphi,\lceil (n_c+1)(1-\alpha) \rceil}) \geq \mathbb{P}(V_{\varphi, n+1} \leq V_{\varphi,\lceil (n_c+1)(1-\alpha) \rceil}) \geq 1-\alpha., \nonumber
\end{equation}
which holds for any $\alpha$, hence $\alpha^*=1$. This concludes statement (i).\\

{\bf (ii) MCX $\boldsymbol{V_\varphi \succeq_{mcx} V^*}$:}

\vspace{.05in} 
By Lemma A4 with $F = F_{V_\varphi}$ and $G = F_{V^*}$, there exists $v^*$ such that $F_{V_\varphi}(v) \leq F_{V^*}(v)$ for all $v\geq v^*$, and $\alpha^* \triangleq 1-F_{V_\varphi}(v^*) \in (0,1]$. Fix $\alpha\in(0,\alpha^*)$. If $\alpha<1/(n_c+1)$, then $Q_{\mathcal{V}_\varphi}(1-\alpha)=\infty$ and coverage holds trivially, so let $k=\lceil (n_c+1)(1-\alpha)\rceil \leq n_c$ and $Q = V_{\varphi,k}$. Conditional on the calibration scores, $Q$ is fixed and $V^*_{n+1}$ and $V_{\varphi,n+1}$ are distributed as $V^*$ and $V_\varphi$, respectively, so that
\[
\mathbb{P}(V^*_{n+1}\leq Q\,|\,\mathcal{D}_c) = F_{V^*}(Q) \geq F_{V_\varphi}(Q)-\mathbf{1}\{Q<v^*\} = \mathbb{P}(V_{\varphi,n+1}\leq Q\,|\,\mathcal{D}_c)-\mathbf{1}\{Q<v^*\}.
\]
Taking expectations and using (\ref{eqA13}) gives $\mathbb{P}(V^*_{n+1}\leq Q) \geq 1-\alpha-\mathbb{P}(Q<v^*)$. The event $\{Q<v^*\}$ occurs only if at least $k$ of the $n_c$ calibration scores are smaller than $v^*$. Their number is Binomial$(n_c, p)$ with $p=\mathbb{P}(V_\varphi<v^*)\leq F_{V_\varphi}(v^*)=1-\alpha^*$, whereas $k/n_c\geq 1-\alpha>1-\alpha^*$. Hence, by Hoeffding's inequality, $\mathbb{P}(Q<v^*)\leq\exp(-2n_c(k/n_c-p)^2)\leq\exp(-2n_c(\alpha^*-\alpha)^2)$. Combining this with (\ref{eqA9})-(\ref{eqA11}) concludes statement (ii). \qed

\renewcommand{\theequation}{A.\arabic{equation}}
\subsection*{A.4. Proof of Theorem 2}

{\bf Theorem 2.} {\it Let $V_{\varphi}(\widehat{\tau}) = |\widehat{\tau}(X)-\widetilde{Y}_{\varphi}|$ and assume that the propensity score function $\pi: \mathcal{X} \to [0, 1]$ is known. Then, the following holds: (i) For the $X$-learner, $V_\varphi$ and $V^*$ do not admit to a model- and distribution-free stochastic order, (ii) For any distribution $P(X, W, Y(0), Y(1))$, CATE estimate $\widehat{\tau}$, and nuisance estimate $\widehat{\varphi}$, the IPW- and the DR-learners satisfy $V_\varphi \succeq_{mcx} V^*$.}

{\it Proof.}\,\, For a CATE estimate $\widehat{\tau}$, the oracle scores are 
\begin{equation}
    V^*(\widehat{\tau}) = |\widehat{\tau}(X)- (Y(1)-Y(0))|.
    \label{eq1Thm2}
\end{equation}
In what follows, we use the notation $V_\varphi(x)$ to denote the conditional random variable $V_\varphi|X=x$, i.e., the conformity score evaluated at a given feature point. Similarly, we use $V^*(x)$ to denote the conditional random variable $V^*|X=x$ (oracle score evaluated at a given feature point).

{\bf Proof of statement (i):}

Recall from Table 1 that the pseudo-outcome for the X-learner is given by:
\begin{equation}
    \widetilde{Y}_{\varphi} = W(Y - \widehat{\mu}_0(X)) + (1-W)(\widehat{\mu}_1(X) - Y).
    \label{eq2Thm2}
\end{equation}
With $V_{\varphi}(\widehat{\tau}) = |\widehat{\tau}(X)-\widetilde{Y}_{\varphi}|$, the conformity score for the X-learner can be written as:
\begin{equation}
    V_{\varphi}(\widehat{\tau}) = |\widehat{\tau}(X)- W(Y - \widehat{\mu}_0(X)) - (1-W)(\widehat{\mu}_1(X) - Y)|.
    \label{eq3Thm2}
\end{equation}
To prove statement (i), it suffices to find counter-examples for estimators $\widehat{\tau}$, $\widehat{\mu}_0$ and $\widehat{\mu}_1$, or data distributions that cause violations of FOSD, SOSD and MCX between $V_\varphi$ and $V^*$.~Let~$u:\mathbb{R} \to \mathbb{R}$~be~a non-decreasing function. The expected value of $u(V_\varphi(x))$ can be evaluated as:
\begin{align}
    \mathbb{E}[u(V_\varphi(x))] &= \mathbb{E}[u(|\widehat{\tau}(x)- W(Y - \widehat{\mu}_0(x)) - (1-W)(\widehat{\mu}_1(x) - Y)|)] \nonumber \\
    &= \pi(x) \cdot \mathbb{E}[u(|\widehat{\tau}(x)- (Y(1) - \widehat{\mu}_0(x))|)] \,\, + \nonumber \\
    &{\color{white}{+}}\,\,(1-\pi(x)) \cdot \mathbb{E}[u(|\widehat{\tau}(x) - (\widehat{\mu}_1(x)-Y(0))|)].
    \label{eq4Thm2}
\end{align}
We start by showing that there exists nuisance and CATE estimates and a data distribution~for~which~$V_\varphi$ does not have FOSD on $V^*$. Assume that $u$ is non-decreasing concave.~By~Jensen~inequality,~we~have:
\begin{align}
    \mathbb{E}[u(V_\varphi(x))] &\leq \mathbb{E}[u(\pi(x) \cdot |\widehat{\tau}(x)- (Y(1) - \widehat{\mu}_0(x))| + (1-\pi(x))\cdot |\widehat{\tau}(x) - (\widehat{\mu}_1(x)-Y(0))|)]. \nonumber 
\end{align}
Now consider a nuisance estimates of $\widehat{\mu}_0=0$ and $\widehat{\mu}_1=0$, a data distribution for which $Y(0)>0$ and $Y(1)<0$ almost surely, and a CATE estimator $\widehat{\tau}(x) > 0,\, \forall x \in \mathcal{X}$, then we have
\begin{align}
    \mathbb{E}[u(V_\varphi(x))] &\leq \mathbb{E}[u(\pi(x) \cdot |\widehat{\tau}(x)- Y(1)| + (1-\pi(x))\cdot |\widehat{\tau}(x) + Y(0)|)], \nonumber \\
    &= \mathbb{E}[u(\pi(x) \cdot (\widehat{\tau}(x)- Y(1)) + (1-\pi(x))\cdot (\widehat{\tau}(x) + Y(0)))], \nonumber \\
    &\leq \mathbb{E}[u(\widehat{\tau}(x) - (Y(1)-Y(0)))],  \nonumber \\
    &\leq \mathbb{E}[u(|\widehat{\tau}(x) - (Y(1)-Y(0))|)] = \mathbb{E}[u(V^*(x))],
    \label{eq5Thm2}
\end{align}
and hence $V^*(x) \succeq_{\mbox{\tiny (2)}} V_\varphi(x)$ for all $x$, and by Lemma A1, we have $V^* \succeq_{\mbox{\tiny (2)}} V_\varphi$.~The~last~inequality in (\ref{eq5Thm2}) holds because $u$ is non-decreasing and $(1-\pi(x))Y(1) - \pi(x) Y(0)$ is always non-positive when $Y(0)>0$ and $Y(1)<0$ (almost surely). Since SOSD is a necessary condition for FOSD, then it follows that $V_\varphi$ does not have FOSD over $V^*$ in this counter-example.  

Now we show that there exists nuisance and CATE estimates and a data distribution~for~which~$V_\varphi$ does not have MCX over $V^*$. This follows directly from the previous counter-example. In the previous counter-example, $V^*(x)-V_\varphi(x) = W\,Y(0)-(1-W)\,Y(1)>0$ almost surely, and hence $\mathbb{E}[V_\varphi]<\mathbb{E}[V^*]$. Since $V_\varphi \succeq_{mcx} V^*$ implies $\mathbb{E}[V_\varphi]\geq\mathbb{E}[V^*]$ (take $u(v)=v$ in Definition 2), $V_\varphi$ does not have MCX over $V^*$.

Finally, we show that there exists nuisance and CATE estimates and a data distribution~for~which~$V^*$ does not have SOSD over $V_\varphi$. Note that $\mathbb{E}[V^*] \geq \mathbb{E}[V_\phi]$ is a necessary condition for $V^* \succeq_{\mbox{\tiny (2)}} V_\varphi$ (See Eq. (4.A.6) in \cite{shaked2007stochastic}). To compare $\mathbb{E}[V^*]$ and $\mathbb{E}[V_\phi]$, we rewrite $\mathbb{E}[V_\phi(x)]$ as:
\begin{align}
    \mathbb{E}[V_\phi(x)] &= \mathbb{E}[|\widehat{\tau}(x)- W(Y - \widehat{\mu}_0(x)) - (1-W)(\widehat{\mu}_1(x) - Y)|], \nonumber \\ 
    &= \pi(x)\cdot \mathbb{E}[|\widehat{\tau}(x)- (Y(1) - \widehat{\mu}_0(x))|] + (1-\pi(x))\cdot \mathbb{E}[|\widehat{\tau}(x)- (\widehat{\mu}_1(x) - Y(0))|], \nonumber \\
    &= \mathbb{E}[\pi(x)\cdot|\widehat{\tau}(x)- (Y(1) - \widehat{\mu}_0(x))| + (1-\pi(x))\cdot |\widehat{\tau}(x)- (\widehat{\mu}_1(x) - Y(0))|], \nonumber \\
    &\geq \mathbb{E}[|\widehat{\tau}(x)- \pi(x)(Y(1) - \widehat{\mu}_0(x))- (1-\pi(x))(\widehat{\mu}_1(x) - Y(0))|], \nonumber \\ 
    &= \mathbb{E}[|\widehat{\tau}(x)- (Y(1)-Y(0)) + (1-\pi(x))(Y(1) - \widehat{\mu}_1(x))+ \pi(x)(\widehat{\mu}_0(x)-Y(0))|]. \nonumber
\end{align}
For nuisance estimates that satisfy $\widehat{\mu}_1(x) < Y(1)$ and $\widehat{\mu}_0(x) > Y(0)$ almost surely, and a CATE estimate that satisfies $\widehat{\tau}(x) \geq Y(1)-Y(0)$ almost surely (e.g., bounded potential outcomes and a sufficiently large constant $\widehat{\tau}$), it follows that $(1-\pi(x))(Y(1) - \widehat{\mu}_1(x))+ \pi(x)(\widehat{\mu}_0(x)-Y(0))$ is always positive and hence we have:
\begin{align}
    \mathbb{E}[V_\phi(x)] &\geq \mathbb{E}[|\widehat{\tau}(x)- (Y(1)-Y(0)) + (1-\pi(x))(Y(1) - \widehat{\mu}_1(x))+ \pi(x)(\widehat{\mu}_0(x)-Y(0))|], \nonumber \\
    &> \mathbb{E}[|\widehat{\tau}(x)- (Y(1)-Y(0))|] = \mathbb{E}[V^*(x)],
\end{align}
which, after averaging over $X$, implies that $V^*$ does not have SOSD over $V_\varphi$.

{\bf Proof of statement (ii):}

From Table 1, the pseudo-outcomes for the IPW- and DR-learners can be written as:
\begin{equation}
    V_{\varphi}(x) = W\,|\widehat{\tau}(x)- Y_\pi(x)| + (1-W)\,|\widehat{\tau}(x)- Y_{1-\pi}(x)|,
    \label{eq6Thm2}
\end{equation}
where $Y_\pi$ and $Y_{1-\pi}$ are defined as follows:
\begin{align}
Y_\pi(x) &= 
    \begin{cases} 
      \frac{1}{\pi(x)}\, Y(1), & \mbox{for IPW-learner}, \\
      \frac{1}{\pi(x)}\, (Y(1)-\widehat{\mu}_1(x)) + (\widehat{\mu}_1(x) - \widehat{\mu}_0(x)), & \mbox{for DR-learner, and}
   \end{cases} \nonumber \\
Y_{1-\pi}(x) &= 
    \begin{cases} 
      \frac{-1}{1-\pi(x)}\, Y(0), & \mbox{for IPW-learner}, \\
      \frac{-1}{1-\pi(x)}\, (Y(0)-\widehat{\mu}_0(x)) + (\widehat{\mu}_1(x) - \widehat{\mu}_0(x)), & \mbox{for DR-learner}.
   \end{cases}
   \label{eq7Thm2}
\end{align}
Note that, following the definition in (\ref{eq7Thm2}), the following holds for both the IPW- and DR-learners:
\begin{align}
\pi(x)\, Y_\pi(x) + (1-\pi(x))\, Y_{1-\pi}(x) = Y(1) - Y(0).
\label{eq8Thm2}
\end{align}
Hence, applying (\ref{eq8Thm2}) to the definition of the oracle score, we can write $V^*(x)$ as follows:
\begin{align}
    V^*(x) = |\widehat{\tau}(x)-(Y(1)-Y(0))| &= |\widehat{\tau}(x)-\pi(x)\, Y_\pi(x) - (1-\pi(x))\, Y_{1-\pi}(x)|, \nonumber \\
    &= |\pi(x)\,(\widehat{\tau}(x)-Y_\pi(x)) + (1-\pi(x))\, (\widehat{\tau}(x)-Y_{1-\pi}(x))|.
    \label{eq9Thm2}
\end{align}
From the expressions in (\ref{eq6Thm2}) and (\ref{eq9Thm2}), we can see that $V_{\varphi}(x)$ is a mixture of the absolute values of the two random variables $(\widehat{\tau}(x)- Y_\pi(x))$ and $(\widehat{\tau}(x)- Y_{1-\pi}(x))$ with mixing proportions $\pi(x)$ and $1-\pi(x)$, and $V^*(x)$ is the absolute value of the convex combination of the same random variables, $(\widehat{\tau}(x)- Y_\pi(x))$ and $(\widehat{\tau}(x)- Y_{1-\pi}(x))$, with weights $\pi(x)$ and $1-\pi(x)$. Applying Lemma A2, it follows that $V_\varphi(x) \succeq_{mcx} V^*(x),\, \forall x \in \mathcal{X}$, and from Lemma A1 we have that $V_\varphi \succeq_{mcx} V^*$.  \qed \\

{\bf Stochastic orders for the signed distance conformity score:}

Throughout, we condition on $\mathcal{D}_\varphi \cup \mathcal{D}_t$, so that $\widehat{\tau}$, $\widehat{\tau}_l$, $\widehat{\tau}_h$, $\widehat{\mu}_0$ and $\widehat{\mu}_1$ are fixed functions, and we assume that $V_\varphi$ and $V^*$ have finite means. Fix $x \in \mathcal{X}$. By unconfoundedness, conditional on $X = x$, the treatment $W \sim \mathrm{Bernoulli}(\pi(x))$ is independent of $(Y(0), Y(1))$, and the pseudo-outcome can be written as $\widetilde{Y}_\varphi = W\,Y_\pi(x) + (1-W)\,Y_{1-\pi}(x)$, where $Y_\pi(x)$ and $Y_{1-\pi}(x)$ are the functions of $(Y(0), Y(1))$ defined in (A.23). Averaging over $W$ first gives, for any function $g$,
\begin{equation*}
\mathbb{E}[g(\widetilde{Y}_\varphi) \mid X = x] = \mathbb{E}\big[\pi(x)\, g(Y_\pi(x)) + (1-\pi(x))\, g(Y_{1-\pi}(x)) \,\big|\, X = x\big].
\end{equation*}
We have shown that for the absolute residual conformity score, $V_{\varphi}(\widehat{\tau}) = |\widehat{\tau}(X)-\widetilde{Y}_{\varphi}|$, the DR- and IPW-learners guarantee $V_\varphi \succeq_{mcx} V^*$ irrespective of the data distribution and underlying models. We now study the stochastic orders achieved by the signed distance conformity score in \cite{linusson2014signed}, defined as: $V_{\varphi}(\widehat{\tau}) = \widehat{\tau}(X)-\widetilde{Y}_{\varphi}$. 
Let $u$ be a non-decreasing concave function. Then, for every $x \in \mathcal{X}$,
\begin{align}
\mathbb{E}[u(V_\varphi(\widehat{\tau})) \mid X = x]
&= \mathbb{E}\big[\pi(x)\, u\big(\widehat{\tau}(x) - Y_\pi(x)\big) + (1-\pi(x))\, u\big(\widehat{\tau}(x) - Y_{1-\pi}(x)\big) \,\big|\, X = x\big] \nonumber\\
&\le \mathbb{E}\big[u\big(\pi(x)\,(\widehat{\tau}(x) - Y_\pi(x)) + (1-\pi(x))\,(\widehat{\tau}(x) - Y_{1-\pi}(x))\big) \,\big|\, X = x\big] \nonumber\\
&= \mathbb{E}\big[u\big(\widehat{\tau}(x) - (Y(1) - Y(0))\big) \,\big|\, X = x\big] \nonumber\\ 
&= \mathbb{E}[u(V^*(\widehat{\tau})) \mid X = x], 
\label{eq10Thm2}
\end{align}
where the inequality applies the concavity of $u$ for each value of $(Y(0), Y(1))$, and the last line follows from (\ref{eq8Thm2}). Since (\ref{eq10Thm2}) holds for every $x \in \mathcal{X}$, taking expectations over $X \sim P_X$ gives $\mathbb{E}[u(V_\varphi(\widehat{\tau}))] \le \mathbb{E}[u(V^*(\widehat{\tau}))]$ for every non-decreasing concave $u$, which is the expected-utility form of SOSD. Hence $V^* \succeq_{(2)} V_\varphi$ (Table 2). Note, however,~that~the~construction~of~predictive intervals with the signed error distance does not follow (\ref{eq7}) since the signed distance score does not sort absolute errors from largest to smallest, but it sorts conformity scores from maximum positive error to maximum negative error. Hence, Theorem 1 does not apply to the signed distance error and the SOSD condition $V^* \succeq_{\mbox{\tiny (2)}} V_\varphi$ does not guarantee coverage. 

Another more commonly used variant of the signed distance score was~proposed~in~\cite{romano2019conformalized}.~Given~a~quantile regression model $[\widehat{\tau}_l, \widehat{\tau}_h]$, \cite{romano2019conformalized} uses a variant of the signed error score defined as:
\begin{align}
V_{\varphi}(\widehat{\tau}) = \max\{\widehat{\tau}_l(X)-\widetilde{Y}_{\varphi}, \widetilde{Y}_{\varphi}-\widehat{\tau}_h(X)\}.
\label{eq11Thm2}
\end{align}
The construction of the predictive intervals in~\cite{romano2019conformalized} follows (\ref{eq7}). Note that $\max\{x, y\}$~can~be~written~as:
\begin{align}
\max\{x, y\} = \frac{x+y}{2} + \frac{|x-y|}{2}. 
\label{eq12Thm2}
\end{align}
Applying (\ref{eq12Thm2}) to (\ref{eq11Thm2}), we can rewrite the conformity scores as:
\begin{align}
V_{\varphi}(\widehat{\tau}) &= \frac{\widehat{\tau}_l - \widehat{\tau}_h}{2} + \frac{|\widehat{\tau}_l + \widehat{\tau}_h - 2 \widetilde{Y}_{\varphi}|}{2}, \nonumber \\
&= \frac{\widehat{\tau}_l - \widehat{\tau}_h}{2} + \left|\frac{\widehat{\tau}_l + \widehat{\tau}_h }{2}-\widetilde{Y}_{\varphi}\right|.
\label{eq13Thm2}
\end{align}
Similarly, the oracle scores can be written as:
\begin{align}
V^*(\widehat{\tau}) &= \frac{\widehat{\tau}_l - \widehat{\tau}_h}{2} + \left|\frac{\widehat{\tau}_l + \widehat{\tau}_h }{2}-(Y(1)-Y(0))\right|.
\label{eq14Thm2}
\end{align}
Let $\delta\tau_-(x) = \frac{\widehat{\tau}_l(x) - \widehat{\tau}_h(x)}{2}$, $\delta\tau_+(x) = \frac{\widehat{\tau}_l(x) + \widehat{\tau}_h(x)}{2}$, and let $u$ be a non-decreasing convex function. Then, for every $x \in \mathcal{X}$,
\begin{align}
\mathbb{E}[u(V_\varphi(\widehat{\tau})) \mid X = x]
&= \mathbb{E}\Big[\pi(x)\, u\big(\delta\tau_-(x) + |\delta\tau_+(x) - Y_\pi(x)|\big) \nonumber\\
&\qquad\;\; + (1-\pi(x))\, u\big(\delta\tau_-(x) + |\delta\tau_+(x) - Y_{1-\pi}(x)|\big) \,\Big|\, X = x\Big] \nonumber\\
&\ge \mathbb{E}\big[u\big(\delta\tau_-(x) + \pi(x)\,|\delta\tau_+(x) - Y_\pi(x)| \nonumber\\ &\qquad\;\; + (1-\pi(x))\,|\delta\tau_+(x) - Y_{1-\pi}(x)|\big) \,\big|\, X = x\big] \nonumber\\
&\ge \mathbb{E}\big[u\big(\delta\tau_-(x) + |\delta\tau_+(x) - \pi(x)\,Y_\pi(x) - (1-\pi(x))\,Y_{1-\pi}(x)|\big) \,\big|\, X = x\big] \nonumber\\
&= \mathbb{E}\big[u\big(\delta\tau_-(x) + |\delta\tau_+(x) - (Y(1) - Y(0))|\big) \,\big|\, X = x\big] \nonumber\\
&= \mathbb{E}[u(V^*(\widehat{\tau})) \mid X = x], 
\label{eq15Thm2}
\end{align}
where the first inequality applies the convexity of $u$ for each value of $(Y(0), Y(1))$ (writing $\delta\tau_-(x) = \pi(x)\,\delta\tau_-(x) + (1-\pi(x))\,\delta\tau_-(x)$), the second follows from the triangle inequality $|\pi a + (1-\pi) b| \le \pi |a| + (1-\pi)|b|$ with $a = \delta\tau_+(x) - Y_\pi(x)$ and $b = \delta\tau_+(x) - Y_{1-\pi}(x)$, together with the monotonicity of $u$, and the last line follows from (\ref{eq8Thm2}) and (\ref{eq14Thm2}). Since (\ref{eq15Thm2}) holds for every $x \in \mathcal{X}$, $V_\varphi(x) \succeq_{mcx} V^*(x)$ for all $x$, and Lemma A1 gives $V_\varphi \succeq_{mcx} V^*$ when the base models apply quantile regression and the conformity scores follow the signed distance score in \cite{romano2019conformalized}. Hence, the guarantee in Theorem 1 applies to conformal meta-learners based on conditional quantile regression.

\subsection*{A.5. Distribution-free validity via superquantile calibration}
Theorem 1 provides no distribution-free lower bound on $\alpha^*$ under condition (ii). To give an example, let $\pi(x)\equiv 1/2$, $Y(1)\equiv 0$ and $Y(0)=sD$, where $s>0$ and $D\sim\mbox{Bernoulli}(p)$ is independent of $X$, with $p\in(0,1/3)$. Then $\mu_1\equiv 0$, $\mu_0\equiv sp$ and $\tau\equiv -sp$. Consider the DR-learner with exact nuisance and CATE estimates, $\widehat{\varphi}=\varphi$ and $\widehat{\tau}=\tau$. Its pseudo-outcome is $\widetilde{Y}_\varphi = -sp$ if $W=1$ and $\widetilde{Y}_\varphi = sp-2sD$ if $W=0$, so that $V_\varphi$ equals $0$, $2sp$ and $2s(1-p)$ with probabilities $1/2$, $(1-p)/2$ and $p/2$, respectively, whereas $V^*=s|D-p|$ equals $sp$ and $s(1-p)$ with probabilities $1-p$ and $p$. By Theorem 2, $V_\varphi\succeq_{mcx}V^*$; the CDFs of $V_\varphi$ and $V^*$ cross finitely many times, with $v^*=s(1-p)$ and $\alpha^*=p/2$. Now fix any $\alpha\in(p/2,p)$. The numbers of calibration scores equal to $0$ and not exceeding $2sp$ are Binomial$(n_c,1/2)$ and Binomial$(n_c,1-p/2)$, respectively, and $1/2<1-\alpha<1-p/2$, so $Q_{\mathcal{V}_\varphi}(1-\alpha)=2sp$ with probability tending to one as $n_c\to\infty$. Since $\mathbb{P}(V^*\leq 2sp)=1-p$, the coverage of $\widehat{C}_\varphi$ converges to $1-p<1-\alpha$. For example, with $p=0.15$, the DR-learner under-covers at $\alpha=0.1$ even with exact nuisance and CATE estimates; the same applies to the IPW-learner. Since, for any $\alpha\in(0,1/3)$, we can choose $p\in(\alpha,\min\{2\alpha,1/3\})$, condition (ii) alone does not guarantee validity at any fixed level of target coverage. 

Our convex dominance result does, however, can be used to control the upper tail of $V^*$ through the {\it superquantile} (i.e., the average of the upper $\alpha$-tail) of $V_\varphi$. Let
\begin{equation}
Q^+_{\mathcal{V}_\varphi}(1-\alpha) \triangleq Q_{\mathcal{V}_\varphi}(1-\alpha) + \frac{1}{\alpha\,|\mathcal{D}_c|}\sum_{k\in\mathcal{D}_c}\big(V_{\varphi,k}-Q_{\mathcal{V}_\varphi}(1-\alpha)\big)_+,
\label{eqSQ}
\end{equation}
and let $\widehat{C}^+_\varphi(x)$ denote the pseudo-interval in (\ref{eq10b}) with $Q_{\mathcal{V}_\varphi}(1-\alpha)$ replaced by $Q^+_{\mathcal{V}_\varphi}(1-\alpha)$.

{\bf Theorem 3.} {\it Let $V_{\varphi}(\widehat{\tau}) = |\widehat{\tau}(X)-\widetilde{Y}_{\varphi}|$ and let $(X_i, W_i, Y_i(0), Y_i(1)),\, i=1,\ldots, n+1$ be i.i.d. If, conditional on $\mathcal{D}_\varphi\cup\mathcal{D}_t$, $V_\varphi\succeq_{mcx}V^*$ and $V_\varphi$ has a continuous distribution with finite mean, then}
\begin{align}
\liminf_{n_c\to\infty}\,\mathbb{P}(Y_{n+1}(1)-Y_{n+1}(0)\in\widehat{C}^+_\varphi(X_{n+1})\,|\,\mathcal{D}_\varphi\cup\mathcal{D}_t) \geq 1-\alpha,\, \forall \alpha \in (0, 1). \nonumber
\end{align}
By Theorem 2, the IPW- and DR-learners satisfy $V_\varphi\succeq_{mcx}V^*$ for any distribution, CATE estimate and nuisance estimate. Hence, whenever their conformity scores are continuous with finite mean, their superquantile-calibrated pseudo-intervals are asymptotically valid at every level of target coverage, with no condition on $\alpha^*$ at the cost of wider intervals, since $Q^+_{\mathcal{V}_\varphi}(1-\alpha)\geq Q_{\mathcal{V}_\varphi}(1-\alpha)$.} 

{\it Proof.}\,\, Throughout, we condition on $\mathcal{D}_\varphi\cup\mathcal{D}_t$, so that $\widehat{\varphi}$ and $\widehat{\tau}$ are fixed, and we consider an i.i.d. sequence of calibration points. The calibration scores $V_{\varphi,1}, V_{\varphi,2},\ldots$ are then i.i.d. copies of $V_\varphi$, independent of $V^*_{n+1}$, which is distributed as $V^*$. Let $F$ be the CDF of $V_\varphi$, $\widehat{F}$ the empirical CDF of the $n_c$ calibration scores, $\mu(t) \triangleq \mathbb{E}[(V_\varphi - t)_+]$, $\mu^*(t)\triangleq\mathbb{E}[(V^*-t)_+]$ and $\widehat{\mu}(t) \triangleq \frac{1}{n_c}\sum_{k\in\mathcal{D}_c}(V_{\varphi,k}-t)_+$, and write $Q = Q_{\mathcal{V}_\varphi}(1-\alpha)$ and $Q^+ = Q^+_{\mathcal{V}_\varphi}(1-\alpha) = Q + \widehat{\mu}(Q)/\alpha$. Since $\{Y_{n+1}(1)-Y_{n+1}(0) \in \widehat{C}^+_\varphi(X_{n+1})\} = \{V^*_{n+1}\leq Q^+\}$, it suffices to show that $\limsup_{n_c\to\infty}\mathbb{P}(V^*_{n+1} > Q^+) \leq \alpha$.

{\it Step 1 (tail bound).} For any $t < q$, we have $\mathbf{1}\{v > q\} \leq (v-t)_+/(q-t)$ for all $v$. Moreover, $v \mapsto (v-t)_+$ is non-decreasing and convex, so $V_\varphi \succeq_{mcx} V^*$ implies $\mu^*(t) \leq \mu(t)$ for all $t$. Hence, on the event $\{\widehat{\mu}(Q) > 0\}$, on which $Q^+ > Q$,
\begin{equation}
\mathbb{P}(V^*_{n+1} > Q^+ \,|\, \mathcal{D}_c) \leq \frac{\mu^*(Q)}{Q^+-Q} \leq \frac{\alpha\,\mu(Q)}{\widehat{\mu}(Q)}.
\tag{$\dagger$}
\label{eqA5dagger}
\end{equation}

{\it Step 2 (consistency).} Let $T \triangleq \inf\{v: F(v)\geq 1-\alpha/2\}$. Since $F$ is continuous, $F(T) = 1-\alpha/2$, so that $\mathbb{P}(V_\varphi > T) = \alpha/2 > 0$ and $\mu(T) > 0$. With $k = \lceil (n_c+1)(1-\alpha)\rceil$, we have $Q \leq T$ if and only if $\widehat{F}(T) \geq k/n_c$. Since $\widehat{F}(T) \to 1-\alpha/2$ almost surely and $k/n_c \to 1-\alpha$, it follows that $0 \leq Q \leq T$ for all sufficiently large $n_c$, almost surely. Next, $\mu$ and $\widehat{\mu}$ are 1-Lipschitz and, by the strong law of large numbers (as $\mathbb{E}[V_\varphi] < \infty$), $\widehat{\mu}(t) \to \mu(t)$ almost surely for every $t$ in any finite grid of $[0, T]$. Together with the Lipschitz property, this gives $\sup_{t \in [0,T]}|\widehat{\mu}(t) - \mu(t)| \to 0$ almost surely. Since $\mu$ is non-increasing, $\mu(Q) \geq \mu(T) > 0$ eventually; therefore, $\widehat{\mu}(Q) > 0$ eventually and $\mu(Q)/\widehat{\mu}(Q) \to 1$ almost surely.

{\it Step 3.} By Steps 1 and 2, $\mathbb{P}(V^*_{n+1} > Q^+\,|\,\mathcal{D}_c) \leq B_{n_c}$, where $B_{n_c} \triangleq \min\{1, \alpha\,\mu(Q)/\widehat{\mu}(Q)\}$ on $\{\widehat{\mu}(Q) > 0\}$ and $B_{n_c} \triangleq 1$ otherwise, and $B_{n_c} \to \alpha$ almost surely. Since $0 \leq B_{n_c} \leq 1$, the reverse Fatou lemma gives $\limsup_{n_c\to\infty}\mathbb{P}(V^*_{n+1} > Q^+) \leq \limsup_{n_c\to\infty}\mathbb{E}[B_{n_c}] \leq \alpha$. \qed

\section*{Appendix B: Literature Review}

In this Section, we provide a comprehensive overview of the relevant~literature.~We~categorize~the previous literature into three distinct strands: (1) Bayesian methods~for~modeling~individualized~causal effects, (2) frequentist methods for inference on parameters pertaining to individualized causal effects, and (3) conformal methods for predictive inference of ITEs.

{\bf (1) Bayesian methods.} Predictive inference on ITEs has been traditionally conducted using Bayesian methods. These methods place a prior distribution over the nuisance parameters $\mu_0$~and~$\mu_1$,~and~then estimate the CATE function through the posterior distribution computed conditional on the observational dataset $\mathcal{D}=\{(X_i, W_i, Y_i)\}_i$. More formally, Bayesian procedures operate as follows: 
\begin{align}
    Y(w) = \mu_w + \epsilon,\,\, \mu_0 \sim \Pi_0,\, \mu_1 \sim \Pi_1, 
\end{align}
where $\Pi_0$ and $\Pi_1$ are priors over function spaces, and $\epsilon \sim \mathcal{N}(0, \sigma^2)$. Given the dataset $\mathcal{D}$,~the~posterior distributions over the two nuisance functions are used to estimate CATE as follows:
\begin{align}
    \widehat{\tau}(x) = \mathbb{E}_{\mu_1 \sim \Pi_1|\mathcal{D}}[\mu_1(x)] - \mathbb{E}_{\mu_0 \sim \Pi_0|\mathcal{D}}[\mu_0(x)].
\end{align}
Furthermore, the posterior distributions $\Pi_0|\mathcal{D}$ and $\Pi_1|\mathcal{D}$ can be used to construct predictive intervals on ITEs through the posterior credible intervals of $P_{\Pi_0}(\mu_1(x)|\mathcal{D})$ and $P_{\Pi_1}(\mu_1(x)|\mathcal{D})$, i.e.,
\begin{align}
    P_{\mu_0, \mu_1\sim \Pi_0, \Pi_1|\mathcal{D}}(Y(1)-Y(0) \in \widehat{C}(X)\,|\,X=x) = 1-\alpha.
\end{align}
Different incarnations of the Bayesian framework correspond to different choices of the prior $(\Pi_0, \Pi_1)$. Bayesian Additive Regression Trees (BART) \cite{chipman2010bart} is one popular model for Bayesian nonparametric regression that was later adapted for causal effect estimation and inference \cite{hill2011bayesian, hahn2020bayesian}. While \cite{hill2011bayesian} applies BART for estimation of CATE and inference of ITE out-of-the-box, \cite{hahn2020bayesian} introduces new regularization techniques to incorporate different forms of inductive biases on the CATE target parameter. Another popular choice of the prior $(\Pi_0, \Pi_1)$ is based on Gaussian processes where~inductive~biases~are~incorporated through different choices of the kernel function of a reproducible~kernel~Hilbert~space~\cite{alaa2017bayesian}. 

Unlike the conformal framework, Bayesian methods do not provide coverage guarantees on their credible intervals. Frequentist coverage can be achieved in an asymptotic sense by Bayesian method under certain technical conditions \cite{sniekers2015adaptive}. Existing methods typically do not provide any finite-sample guarantees and the achieved empirical coverage depends on the choice of the prior. Models such as BART are typically applied with a default prior for all datasets, undermining their achieved coverage in diverse experimental setup. In \cite{lei2021conformal}, it was shown that Bayesian methods have~a~tendency~to~undercover ITEs in some data generation processes (e.g., in high-dimensional covariate spaces). 

Bayesian methods also suffer from two key advantages compared to the conformal framework. First, Bayesian inference procedures and the computation of posterior distributions are typically tailored to specific model architectures. Generalizing exact Bayesian inference to arbitrary model architectures, including modern deep learning architectures, is not straightforward. Moreover, Bayesian inference is conducted either using computationally exhaustive Monte Carlo method (e.g., BART) or through expensive evaluation of an analytical posterior distribution that does not scale well with the number of training points (e.g., Gaussian process posteriors are $\mathcal{O}(n^3)$). 

{\bf (2) Frequentist methods.} Another model-specific approach to inference on causal parameters is the Causal Forest model in \cite{wager2018estimation}. This procedure uses constructs pointwise confidence intervals on the CATE function $\tau(x)$ using conditional variance estimates based on the infinitesimal jackknife \cite{wager2014confidence}. These intervals provide asymptotically valid coverage of CATE under mild regularity assumptions, but they provide no (asymptotic or finite-sample) guarantees on coverage for ITEs. 

Another non-Bayesian approach for predictive inference of ITEs uses~a~quantile~regression~approach~to train models that construct upper and lower bounds on the observed potential outcomes using a pinball loss \cite{makar2020estimation}. While quantile regression can be repurposed for different~model~architectures,~this~approach does not provide finite-sample coverage guarantees and is empirically~found~to~undercover~outcomes, hence it is typically supplemented with a conformal prediction procedure \cite{romano2019conformalized}.

{\bf (3) Conformal methods.} The application of conformal prediction to ITE inference traces its origins back to the covariate shift problem. In \cite{tibshirani2019conformal}, a variant of conformal prediction was proposed to address problem setups where training and testing data are drawn from different distributions, breaking the exchangeability assumptions required for valid inference. 

In the treatment effect estimation setup, covariate shift arises because the distributions of the treatment groups differ from that of the target population. Recall from (\ref{eq8c}) that the conformity scores for the nuisance estimates $\widehat{\mu}_0$ and $\widehat{\mu}_1$ can be written as follows:
\begin{equation} 
\colorbox{blue!5}{$V^{(0)}_k(\widehat{\mu}_0) \triangleq V(X_k, Y_k(0); \widehat{\mu}_0),\, \forall k \in \mathcal{D}_{c,0},$}\,\,\,\,\,\,\,\, \colorbox{red!10}{$V^{(1)}_k(\widehat{\mu}_1) \triangleq V(X_k, Y_k(1); \widehat{\mu}_1),\, \forall k \in \mathcal{D}_{c,1},$} \nonumber
\end{equation}
Both sets of conformity scores are sampled from the populations $P_{X|W=0}$ and $P_{X|W=1}$, respectively, and hence they are not exchangeable with the conformity scores sampled from target covariate distribution $P_X$. To retain the validity of intervals constructed using the sample quantiles of conformity scores, \cite{tibshirani2019conformal} introduces the notion of ``weighted exchangeability'' and proposes a weighted sample quantile with weights based on the likelihood ratios between training and testing distributions. \cite{lei2021conformal} uses this method to construct valid intervals for potential outcomes and then proposes different approaches for combining these intervals to construct intervals for ITEs. Other work utilized the conformal prediction framework to conduct sensitivity analysis by devising hypothesis tests for the presence of hidden confounders \cite{jin2023sensitivity, yin2022conformal}. In \cite{zhang2023conformal, taufiq2022conformal}, conformal prediction has been utilized to address the closely related problem of off-policy evaluation. More recently, there has been further work developing methods for conformal prediction under covariate shift \cite{gibbs2021adaptive, barber2022conformal, qiu2022distribution}. 

The key difference between our work and the aforementioned approaches is that we do not apply conformal prediction to the nuisance parameters, and instead conduct inference directly on the target parameter. We do so by applying conformalization with respect to transformed targets \mbox{\small $\widetilde{Y}_\varphi$} for each data point in $\mathcal{D}$, and this way the covariate shift problem does not arise since \mbox{\small $(X, \widetilde{Y}_\varphi)$} is sampled from the same distribution in calibration and testing data. One way to think about our approach as opposed to weighted conformal prediction (WCP) is that ours applies re-weighting to the outcomes whereas WCP re-weights the conformity scores. By moving the re-weighting step to the outcomes, we decouple the conformal procedure from the model, allowing for more flexible choice of inductive priors and accommodating models that do not estimate potential outcomes directly.

\section*{Appendix C: Experimental Setup and Results}

\subsection*{C.1. Overview of the experimental setup}

In all experiments, we used a Gradient Boosting model with 100 trees as the base model for nuisance estimation and quantile regression on pseudo-outcomes. We used the same base model in conformal meta-learners as well as the weighted CP (WCP) baselines to control for potential~performance~differences resulting from different choices of base models. Unless otherwise stated, all experiments followed a 90\%/10\% train/test split of each dataset, and each training set with further split into a 75\%/25\% proper training/calibration sets. All performance metrics (empirical coverage, expected interval width and RMSE) are evaluated on the testing data and averaged over 100 runs. The target coverage in all experiments was set $1-\alpha=0.9$.

We used the official implementations of the (na\"ive, exact and inexact) WCP baselines in the R package available at \href{https://github.com/lihualei71/cfcausal}{https://github.com/lihualei71/cfcausal}. We executed these baselines through rpy2 (\href{https://rpy2.github.io/}{https://rpy2.github.io/}) wrappers within our Python codebase. 

\subsection*{C.2. Details of the IHDP and NLSM datasets}

In this Section, we provide details of the semi-synthetic datasets used in the experiments in Section 5. 

{\bf IHDP dataset.} The Infant Health and Development Program (IHDP) was a multi-site randomized controlled trial that studied the effect of early educational intervention on the cognitive development of low birth weight premature infants \cite{brooks1992effects}. The trial targeted low-birth-weight, premature infants, and administered an intensive high-quality child care and home visits from a trained provider to the treatment group. In \cite{hill2011bayesian}, a semi-synthetic benchmark was developed based on the covariate data in IHDP with simulated outcomes to assess the performance of different CATE estimation and ITE inference models when ground-truth counterfactual are known. The dataset comprised 25 covariates (6 continuous and 19 binary variables) capturing aspects related to children and their mothers. The benchmark dataset excluded a non-random subset of treated individuals \cite{hill2011bayesian}. The final dataset consists of 747 samples (139 treated and 608 control). The outcomes of all individuals were simulated whereas the treatment assignments followed the true assignments in the study.

The simulated outcomes model in IHDP followed simulation Setup B in \cite{hill2011bayesian}.~In~this~Setup,~the~potential outcomes were simulated as follows: $Y(0) \sim \mathcal{N}(\exp((X + W)\beta), 1)$ and $Y(1) \sim \mathcal{N}(\exp(X \beta) - \omega, 1)$, where $W$ is an offset matrix, $\omega$ is set so that the ATE on the treated is always equal to 4, and the coefficients $\beta$ are sampled independently at random from $(0, 0.1, 0.2, 0.3, 0.4)$ with probabilities $(0.6, 0.1, 0.1, 0.1, 0.1)$. In our experiments, we used the 100 realization of~the~training and testing data released by \cite{shalit2017estimating} in \href{https://www.fredjo.com/files/ihdp_npci_1-100.train.npz}{https://www.fredjo.com/files/ihdp\_npci\_1-100.train.npz} and \href{https://www.fredjo.com/files/ihdp_npci_1-100.test.npz}{https://www.fredjo.com/files/ihdp\_npci\_1-100.test.npz}.

{\bf NLSM dataset.} The National Study of Learning Mindsets (NLSM) is a large-scale randomized trial that studied the effect of a behavioral intervention, designed to instill students~with~a~growth~mindset, on the academic performance of students in secondary education in the US \cite{yeager2019national}. In \cite{carvalho2019assessing}, a semi-synthetic benchmark was developed based on the covariates of the NLSM study. Unlike IHDP, this benchmark does not use the real covariates of NLSM but rather buids a synthetic process to emulate NSLM in terms of the covariate distributions, data structures, and effect sizes.~The~final~dataset~comprised 8,132 data points. The data was generated based on 5 principles (See Section 2 and Appendix A in \cite{carvalho2019assessing}). The principles are: (1) ATEs should be well-estimated from synthetic~data~by~any~reasonable procedure, (2) variability in CATE should be relatively modest, (3) treatment effect heterogeneity can be approximately recovered given complete knowledge of the correct functional form, (4) no additional unmeasured treatment effect moderation at the individual level,~and~(5)~unexplained~treatment effect heterogeneity at the group level should be present at reasonable levels.

\section*{Errata}

This version corrects the following errors and gaps in the theoretical results and experiments of the previous version. These errors were identified and corrected with the assistance of Claude Opus 5.5 (Anthropic).

\begin{enumerate}
\item \textbf{Second-order dominance in Theorem~1.} The previous version listed $V_\varphi \preceq_{(2)} V^*$ as a sufficient  condition for validity. It is not sufficient in general: it holds when $V_\varphi \equiv 0$ and $V^* \equiv 1$, where coverage is zero. When $V_\varphi$ and $V^*$ have equal means, it is equivalent to the convex order, which is covered by the corrected convex dominance condition (now condition~(ii)). Thus, this condition has been removed.

\item \textbf{Convex dominance in Theorem~1.} The previous proof relied on a corollary stating that the CDFs of $V_\varphi$ and $V^*$ cross at a single point. Convex dominance does not generally imply this, and the CDFs can cross several times (Appendix~A.5). The corrected proof uses only the last crossing point $v^*$. The new Lemma~A4 shows that if $F_{V_\varphi} - F_{V^*}$ changes sign finitely many times, then $F_{V_\varphi} \le F_{V^*}$ beyond $v^*$ and $\alpha^* = 1 - F_{V_\varphi}(v^*) > 0$; this condition was added to Theorem~1. The corrected theorem also accounts for the calibration quantile falling below $v^*$ in finite samples through a slack term $\exp(-2n_c(\alpha^* - \alpha)^2)$. 

\item \textbf{New result.} Since convex dominance does not keep $\alpha^*$ away from zero, we provide a new result in Theorem~3 showing that calibrating with the superquantile of the pseudo-outcome scores gives asymptotically valid coverage at every level under convex dominance alone.

\item \textbf{Experiments.} We found and fixed the following errors in the experiment code, and re-ran all experiments in Section~5: (a) the RMSE was averaged over all pairs of test units rather than computed unit by unit, i.e. the classic Python error of multiplying (n, 1) and (n,) arrays, which inflated the RMSE of all methods whenever the CATE varies across units (i.e., in all experiments except \textbf{Setup A}); (b) the X-learner used the plug-in pseudo-outcome $\widehat{\mu}_1(X) - \widehat{\mu}_0(X)$ instead of the pseudo-outcome in Table~1; (c) in IHDP, the potential outcomes used for evaluation were rescaled by the standard deviation of the CATE (when larger than one), but the factual outcomes used for training and calibration were not; and (d) in NLSM, five of the eleven covariates were dropped from all models. The coverage results and the empirical stochastic orders are qualitatively unchanged: the IPW- and DR-learners achieve the target coverage in all experiments, and the X-learner under-covers. With the corrected RMSE, however, the WCP baselines provide more accurate point estimates than the DR-learner in \textbf{Setup B}, IHDP and NLSM, and Section~5.2 was revised accordingly. 
\end{enumerate}

\end{document}